\documentclass{article}

\usepackage{microtype}
\usepackage{graphicx}
\usepackage{subcaption}
\usepackage{booktabs} 

\usepackage{multirow}
\usepackage{placeins}

\usepackage{hyperref}

\usepackage[accepted]{mlsys2025}

\usepackage{enumitem}
\setlist{nosep}

    \setlist[itemize]{leftmargin=10pt, itemsep=0pt, parsep=0pt}

\mlsystitlerunning{Measuring the Effectiveness of RLVR in Low Data and Compute Regimes}

\begin{document}

\twocolumn[
\mlsystitle{Learning from Less: Measuring the Effectiveness of RLVR in Low Data and Compute Regimes}



\mlsyssetsymbol{nowreflection}{*}
\mlsyssetsymbol{nowoxford}{$\dagger$}

\begin{mlsysauthorlist}
\mlsysauthor{Justin Bauer}{snorkel}
\mlsysauthor{Thomas Walshe}{snorkel,nowreflection}
\mlsysauthor{Derek Pham}{snorkel}
\mlsysauthor{Harit Vishwakarma}{snorkel,nowoxford}
\\
\mlsysauthor{Armin Parchami}{snorkel}
\mlsysauthor{Frederic Sala}{snorkel,wis}
\mlsysauthor{Paroma Varma}{snorkel}
\end{mlsysauthorlist}

\mlsysaffiliation{snorkel}{Snorkel AI}
\mlsysaffiliation{wis}{University of Wisconsin-Madison}

\mlsyscorrespondingauthor{Justin Bauer}{justin.bauer@snorkel.ai}

\mlsyskeywords{Machine Learning, MLSys}

\vskip 0.3in

\begin{abstract}
Fine-tuning Large Language Models (LLMs) typically relies on large quantities of high-quality annotated data, or questions with well-defined ground truth answers in the case of Reinforcement Learning with Verifiable Rewards (RLVR). While previous work has explored the benefits to model reasoning capabilities by scaling both data and compute used for RLVR, these results lack applicability in many real-world settings where annotated data and accessible compute may be scarce. In this work, we present a comprehensive empirical study of open-source Small Language Model (SLM) performance after RLVR in low data regimes. Across three novel datasets covering number counting problems, graph reasoning, and spatial reasoning, we characterize how model performance scales with dataset size, diversity, and complexity. We demonstrate that (1) procedural datasets allow for fine-grained evaluation and training dataset development with controllable properties (size, diversity, and complexity), (2) under RLVR, models trained on lower complexity tasks can generalize to higher complexity tasks, and (3) training on mixed complexity datasets is associated with the greatest benefits in low data regimes, providing up to 5$\times$ sample efficiency versus training on easy tasks. These findings inspire future work on the development of data scaling laws for RLVR and the use of procedural data generators to further understand effective data development for efficient LLM fine-tuning. 
\end{abstract}
]



\printAffiliationsAndNotice{$^*$Work done at Snorkel AI. Now at Reflection AI. $^\dagger$Work done at Snorkel AI. Now at University of Oxford.}

\section{Introduction}
Recent advances in Large Language Models (LLMs) have achieved significant improvements in reasoning capabilities~\cite{openai2025gpt5, comanici2025gemini, openai2025o3_n_o4mini, zeng2025glm, anthropic2025claudeSonnet45}; this has, in part, been driven by the adoption of Reinforcement Learning with Verifiable Rewards (RLVR)~\cite{shao2024deepseekmath, wen2025reinforcement}. RLVR provides an effective method for post-training LLMs by rewarding models based on verifiable outcomes (e.g., answers that can be compared to a known unambiguously correct ground truth) rather than noisy human preferences~\cite{poddar2024personalizing}. For classes of problems with verifiable outcomes, such as in mathematics~\cite{wang2025reinforcement}, the adoption of RLVR has enabled models (e.g., DeepSeek R1~\cite{guo2025deepseek}) to achieve state-of-the-art performance and allowed strong problem solving and self-correction capabilities to emerge. However, many of these advances are made under the assumption that high-quality training data and compute are abundant~\cite{khatri2025art}.

Recent RLVR research has focused on using high volumes of question-answer pairs to improve the reasoning capabilities of LLMs. For example, DeepMath-103K includes over 100,000 challenging and decontaminated samples for training~\cite{he2025deepmath}. However, in realistic, resource-constrained situations, where both annotated data and compute may be limited, these results may be challenging to replicate or extend to new reasoning domains. While previous studies have explored scaling RLVR with respect to model size and compute budget \cite{khatri2025art, tan2025scaling}, or focused on reducing compute requirements (both through Small Language Model (SLM) fine-tuning~\cite{dang2025reinforcement} and Low-Rank Adaptation (LoRA)~\cite{wang2025tina}), there has been less attention on data scaling and the effectiveness of RLVR in low data regimes. In this work, we focus on characterizing how the size, diversity, and complexity of training data influence the reasoning capabilities of models. Motivated by these ideas, we specifically investigate the research question: \textit{``How does model performance evolve when training data and compute are limited, and what characteristics of data impact generalization in such regimes?''}

We conduct a systematic empirical study using open-source SLMs fine-tuned using RLVR under low data and compute regimes to help understand these relationships, the results of which help to shape data scaling laws that describe the effectiveness of RLVR in different scenarios. For our experiments, we introduce three novel datasets that allow data to be generated procedurally with desired volume, diversity, and complexity. These procedural datasets allow us to better isolate the influence of different attributes (e.g., topic of questions, complexity of questions, etc.) and study fine-tuning dynamics in controlled settings. We summarize our contributions and findings as follows:

\begin{itemize}
    \item \textbf{New procedural datasets for reasoning tasks.} We develop three new datasets designed to support RLVR that cover number counting problems, graph reasoning, and spatial reasoning. For each of the datasets, we report empirical solving rates across 10 models (including open-source and proprietary LLMs), demonstrating the value of using procedural data in creating challenging and complex tasks.
    \item \textbf{Study of RLVR in low data regimes.} Using an open-source LLM (Qwen3-4B), we investigate the relationship between dataset size and composition (focusing on task complexity) and performance after fine-tuning. Different training data configurations are used, capturing a range of sizes and complexities across the three dataset types. We observe that models trained on small volumes of lower complexity tasks (i.e., easier questions) generalize to more complex tasks, and that training on a mixed complexity dataset is associated with up to 5$\times$ the sample efficiency under the same data budget.
\end{itemize}


\section{Related Work}

\subsection{Scaling Laws for Language Models}

Early work by \citet{kaplan2020scaling} and \citet{hoffmann2022training} established predictable relationships between model performance, size, data, and compute, showing that increasing training data and parameters yields smooth performance improvements under fixed budgets. These analyses primarily characterize how performance scales with model size and compute, but do not address the effect of dataset composition, particularly difficulty distribution, under fixed budgets. \citet{zhang2024scaling} extended these ideas to supervised fine-tuning, showing that downstream loss depends jointly on fine-tuning data size, model size, and parameter-efficient adaptation methods.

\subsection{Reinforcement Learning Scaling for LLMs}

Recent work has explored how RL affects post-training performance in LLMs. \citet{khatri2025art} present ScaleRL, a large-scale framework characterizing RL performance under different compute budgets and algorithmic choices, providing valuable insights into efficiency at scale. However, their study primarily emphasizes compute scaling rather than data composition or limited-data regimes. \citet{tan2025scaling} analyze RL post-training in mathematical reasoning tasks using Qwen2.5 models (0.5B--14B), finding that larger models achieve higher sample efficiency and that moderate data reuse ($\leq$25 epochs) approaches the performance of unique data. However, their analysis is confined to a single domain and does not explore dataset composition or difficulty effects. \citet{lai2025survey} provide a comprehensive overview of post-training methods, including supervised fine-tuning and RL from feedback. Their taxonomy outlines general scaling trends but remains largely descriptive, underscoring the need for empirical studies focused on data composition rather than model or compute scaling.

\subsection{Data Efficiency and Selection in RL}

Efficient data utilization has been a recurring challenge in RL fine-tuning. \citet{shen2025exploring} identify two major bottlenecks in RLHF data scaling---reward hacking and reduced response diversity---and propose hybrid reward systems and prompt-selection strategies that emphasize harder, low-reward prompts to stabilize training and improve reasoning. Their focus on selecting informative examples aligns with our investigation into how dataset composition and difficulty influence RL performance under limited data. \citet{li2024limr} introduce LIMR (Less is More for RL), showing that small amounts of carefully curated data can outperform larger datasets. By quantifying each sample's contribution to learning, they demonstrate that 1.4K selected samples can match the performance of 8.5K unfiltered ones on mathematical reasoning tasks. While they provide a selection heuristic, they do not analyze how varying dataset size or difficulty affects performance, which our study examines.

\subsection{Verification and Reasoning}

Work on verifiable rewards has advanced understanding of how structured feedback can drive reasoning improvements. \citet{wen2025reinforcement} demonstrate that RLVR expands genuine reasoning ability rather than simply improving sampling efficiency, introducing CoT-Pass@K metrics to measure both reasoning and answer correctness. \citet{liu2025trust} propose RISE, an online RLVR framework that jointly optimizes problem-solving and self-verification, improving verification accuracy and test-time robustness. These insights inform our reward design, which combines correctness verification with format and conciseness components to provide denser feedback in low-data training settings.

Our study complements these lines of research by holding model size and compute fixed and isolating the effect of data composition on RLVR effectiveness, motivating future work on budget-aware RLVR theory that captures interactions between optimization budget, token limits, and reward sparsity.

\section{Methodology}

\subsection{Datasets}
\label{sec:datasets}

To evaluate model performance with controlled training data, we designed three programmatically generated datasets: (1) Counting Problems, (2) Graph Reasoning, and (3) Spatial Reasoning. Each dataset was constructed using pre-defined code templates that allow for parameterization and controlled variation across a well-defined taxonomy of operators, ranges, and/or conditions. All data instances contain verifiable outcome-level ground truth, enabling both quantitative evaluation and use in RLVR training pipelines without relying on costly annotation or less reliable verification methods (e.g., LLM-as-judge evaluation). The dataset taxonomies are hand-crafted to promote diversity through meaningful variation rather than to provide statistical guarantees of coverage. Across all three datasets, the procedural generators control multiple correlated instance properties simultaneously (e.g., graph size and edge density, operator families and step counts, number of actions and query types), so comparisons between Easy and Mixed training configurations reflect variation along several dimensions rather than a single axis of complexity. 

\subsubsection{Counting Problems Dataset}

The Counting Problems dataset is a procedurally generated benchmark designed to evaluate the numerical reasoning and pattern recognition capabilities of language models in constrained computational tasks. The dataset employs a templating methodology to generate questions with deterministic programmatic ground-truth answers, enabling precise control over problem complexity and systematic evaluation of model performance. Each question is composed of:
\begin{itemize}
    \item A natural language prompt that specifies a counting task over a sequence of integers within a defined range.
    \item A sequence of conditional filters and transformations to apply before the final counting operation.
    \item A deterministic ground-truth answer computed by programmatically executing the specified operations.
\end{itemize}

This design allows for direct assessment of multi-step reasoning and numerical manipulation, as models must accurately parse operation sequences, track intermediate states, and perform multi-step reasoning to produce correct answers.

\textbf{Controllable Complexity.}
We control problem complexity through multiple structural dimensions. Primarily, we vary the \textit{range scale}, which controls the magnitude of integer ranges from which values are drawn. We also manipulate the \textit{operator diversity} through a taxonomy of counting and aggregation operations:
\begin{itemize}
    \item \textbf{Basic Counting:} Count, Unique Count, Zero Count.
    \item \textbf{Conditional Counting:} Even Count, Odd Count, Positive Count, Negative Count, Divisible By N Count.
    \item \textbf{Threshold-Based:} Below Threshold Count, Above Threshold Count.
    \item \textbf{Arithmetic Aggregation:} Sum, Product, Mean, Median, Mode.
    \item \textbf{Extrema Operations:} Min, Max, Range.
    \item \textbf{Bitwise Operations:} Bitwise AND, Bitwise OR, Bitwise XOR, Bitwise NAND.
\end{itemize}
Additionally, we vary \textit{compositional depth} through the number of conditional filters (1--4) and transformations (0--3) applied before the final operation, creating problems with 1--7 total intermediate steps. This yields a spectrum from simple counting to complex multi-step compositional reasoning.

\textbf{Example.} The following is an example counting problem.

\textbf{Question:} Consider the integers from 1 to 100, inclusive. First, keep only the numbers that are even. Then, keep only the numbers that are divisible by 3. Of these numbers, count how many values remain.\\
\textbf{Solution:} 16 (computed programmatically by executing: $|\{x \in [1,100] : x \bmod 2 = 0 \land x \bmod 3 = 0\}| = 16$)

\textbf{Evaluation Protocol.}
Each model receives one prompt per question and generates a completion containing reasoning (optional) and a final numerical answer. Numeric responses are parsed using regular expressions and programmatically validated via exact-match comparison against the deterministic ground truth.

\subsubsection{Graph Reasoning Dataset}

The Graph Reasoning dataset is a procedurally generated benchmark designed to evaluate the mathematical and spatial reasoning capabilities of language models over graph-structured problems. The dataset extends the templating methodology to formal graph-based domains, allowing precise control over question complexity and solution verifiability. Each question is composed of:
\begin{itemize}
    \item A natural language operator that defines a computation over a graph (e.g., ``Find the minimum vertex cover of an undirected graph'').
    \item A graph structure encoded textually as lists of nodes and edges.
    \item A verifiable ground-truth solution, computed via deterministic algorithms.
\end{itemize}

This design allows for direct assessment of multi-hop reasoning and long-context tracking, as models must parse graph representations and perform symbolic reasoning to produce correct answers. 

\textbf{Controllable Complexity.} 
We control problem complexity through several factors. Primarily, we vary the \textit{graph size} (number of nodes and edges), ranging from small (5 nodes) to large (25 nodes) graphs, which increases the reasoning load and relational tracking required. We also vary the \textit{operator diversity}, drawing from a predefined taxonomy of graph-theoretic operations that span multiple problem families:
\begin{itemize}
    \item \textbf{Subgraph Optimization:} Minimum Density Subgraph, Maximum Clique, Maximum Independent Set, Minimum Vertex Cover, Maximum Induced Bipartite Subgraph, Acyclic Subgraph, Dense Subgraph Variants.
    \item \textbf{Graph Partitioning:} Balanced Cut.
    \item \textbf{Feedback Set Problems:} Feedback Vertex Set, Feedback Edge Set.
    \item \textbf{Path Problems:} Longest Path, Hamiltonian Path, Hamiltonian Cycle.
    \item \textbf{Graph Metrics:} Graph Diameter, Graph Radius, Graph Density.
\end{itemize}
Additionally, some instances include \textit{weighted} and \textit{directed} edges to introduce further structural variation, though these are not treated as primary experimental variables.

\textbf{Example.} The following is an example question.

\textbf{Question:} Find the maximum independent set of an undirected graph with 5 nodes. Find the largest set of vertices with no edges between them. If multiple maximum independent sets exist, return any one of them.\\
\textbf{Graph:} Nodes: [0, 1, 2, 3, 4]; 
Edges: [(0,2), (0,4)].\\
\textbf{Solution:} The maximum independent set is [1, 2, 3, 4].

\textbf{Evaluation Protocol.}  
Each model receives one prompt per question and is required to generate both a complete reasoning trace and a final answer. The responses are subsequently normalized by a secondary model (GPT-4o) into a canonical internal representation. A programmatic validator, implemented using graph-based libraries (e.g., \texttt{networkx}), then verifies each output for correctness against the ground truth or determines whether it constitutes a valid equivalent solution.

\subsubsection{Spatial Reasoning Dataset}
The Spatial Reasoning dataset evaluates spatial reasoning capabilities of language models. We generate the problems with varying difficulty level following the spatial reasoning setup introduced in \cite{dsouza2025automatingbenchmarkdesign}. Similar to the graph reasoning and counting problems, this setting extends the templating methodology and allows precise control over question complexity and solution verifiability. Each question is composed of:
\begin{itemize}
    \item A description of a 2D spatial reasoning environment consisting of a square grid (board) and a set of particles on the board. The board and particles are located in a 2D space and are oriented towards one of the cardinal directions (East, North, West, South).
    \item A sequence of movement and rotation actions applied to the board and particles.
    \item A query about the absolute or relative location or orientation of the entities (board or particles) after all actions have been applied.
    \item A verifiable ground-truth solution, obtained by programmatically executing the simulation.
\end{itemize} 

These problems test LLMs' abilities to track and reason over the location and orientation of entities in 2D space. The division of problems between absolute and relative is motivated by the fundamental dichotomy of egocentric (relative) and allocentric (absolute) spatial reasoning depending on the frame of reference \cite{denis2017space}. 

\textbf{Controllable Complexity.} We control the complexity via the number of actions and the type of query. Intuitively, the problems with more actions and the ones based on the relative spatial reasoning are expected to be more complex.

\textbf{Example.} The following is a sample problem.

\textbf{Question:} Consider a square grid of size $20\times20$ centered at $(0,0)$. It has two particles P1 and P2 at locations $(-1.5, 2.5)$ and $(3.5, 1.5)$, respectively. P1 and P2 face towards East and West, respectively. P1 moves 1 step forward and P2 moves 1 step backwards. What is the location of P1, relative to P2?

\textbf{Solution:} The location of P1 relative to P2 is $(-5.0, 1.0)$.

\textbf{Evaluation Protocol.}
Models are prompted with the question and asked to generate a completion containing an optional explanation and answer in structured (JSON) format. If the original response has parsing errors, we fall back to parsing with a secondary model (GPT-4o). The structured response is then compared against the ground truth. For floating-point numbers, we match to the first 3 decimal places and perform an exact match for integer and string values.  

\subsection{Curation}
\label{sec:curation}

For our model training experiments, we generated over 1,500 programmatically defined problems for each dataset described above. Following dataset generation, we conducted model-based evaluation runs based on each dataset's evaluation protocol across 10 diverse foundation models spanning several model families (GPT, Claude, Gemini, Grok, Llama, and Qwen).
\begin{figure}[t]
    \centering
    \includegraphics[width=0.95\linewidth]{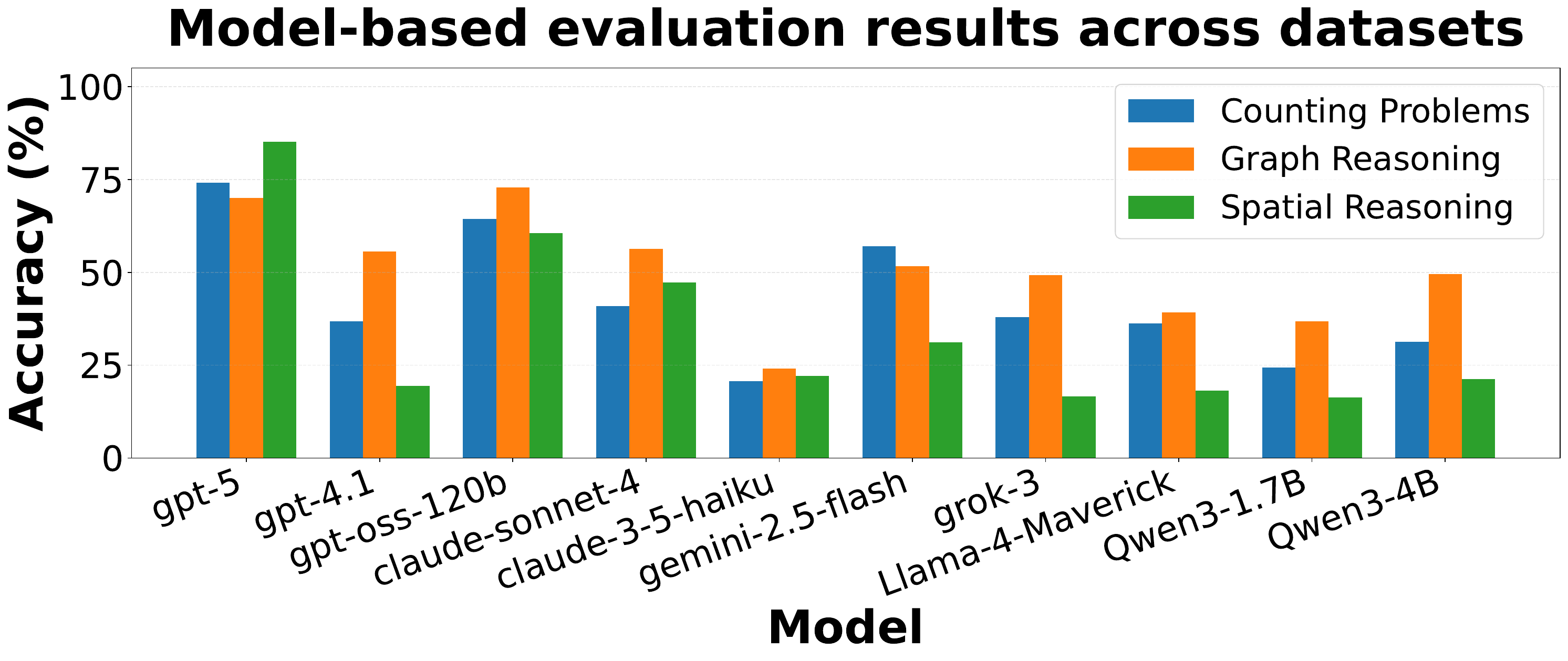}
    \caption{Overall model-based evaluation results across all generated samples for Counting Problems, Graph Reasoning, and Spatial Reasoning.}
  \label{fig:eval_performance}
\end{figure}

Each model was evaluated with a single inference call per data point, and aggregate pass rates were computed at the instance level. Figure~\ref{fig:eval_performance} shows each dataset's performance across the ten LLMs we used. To enable finer-grained difficulty control, we categorized all problems into three difficulty tiers based on the percentage of models that answered correctly:
\begin{itemize}
    \item \textbf{Easy:} 67--100\% of models answered correctly.
    \item \textbf{Medium:} 34--66\% of models answered correctly.
    \item \textbf{Hard:} 0--33\% of models answered correctly.
\end{itemize}

We then curated multiple subsets for downstream use by sampling based on difficulty. The following dataset configurations were curated:
\begin{itemize}
    \item \textbf{Easy training subsets:} 100, 200, and 500 examples sampled from the Easy tier.
    \item \textbf{Mixed training subsets:} 100, 200, and 500 examples drawn across Easy, Medium, and Hard tiers (approximately 33\% each).
    \item \textbf{Test subset:} 200 examples (500 examples for Graph Reasoning), not used in training, drawn across Easy, Medium, and Hard tiers.
\end{itemize}

This multi-model calibration and stratified curation distinguishes difficulty based on performance across a diverse range of models, ensuring that difficulty labels reflect actual model capabilities across architectures rather than human assumptions. Because multiple instance properties covary with difficulty tier, observed performance differences between Easy and Mixed configurations should not be attributed to a single factor. The result is a set of practical datasets that capture realistic capability boundaries across the current LLM frontier, enabling controlled studies of how problem difficulty interacts with fine-tuning dataset scale. All training, validation, and test splits are strictly disjoint to prevent data contamination or leakage between stages.

\subsection{Fine-tuning}
\label{sec:reinforce}

We employ RLVR to fine-tune SLMs on each problem domain. This approach enables direct optimization of task-specific reward signals rather than relying on supervised demonstrations, allowing the model to learn through exploration and self-correction.

\textbf{Base Model and Architecture.}
We use \texttt{Qwen3-4B}~\citep{qwen3} as our base model across all experiments, a 4-billion parameter model with strong reasoning capabilities. To enable efficient fine-tuning on consumer hardware, we apply LoRA~\citep{hu2021lora} with rank $r=64$ and $\alpha=16$, targeting all linear layers. Recent work shows LoRA matches full fine-tuning performance for reinforcement learning even at low ranks~\citep{schulman2025lora}, supporting its use in our compute-constrained setting. This reduces trainable parameters to $\sim$100M while preserving model expressiveness.

\begin{table}[b]
\centering
\caption{Reinforcement learning hyperparameters across datasets.}
\label{tab:rl_hyperparameters}
\small
\resizebox{\columnwidth}{!}{%
\begin{tabular}{lrrr}
\toprule
\textbf{Hyperparameter} & \textbf{Counting} & \textbf{Graph Reasoning} & \textbf{Spatial Reasoning} \\
\midrule
\multicolumn{4}{l}{\textit{Shared Architecture Parameters}} \\
Base Model & \multicolumn{3}{c}{Qwen3-4B} \\
LoRA Rank & \multicolumn{3}{c}{64} \\
LoRA Alpha & \multicolumn{3}{c}{16} \\
\midrule
\multicolumn{4}{l}{\textit{Training Parameters}} \\
Training Steps & 300 & 300 & 1000 \\
Learning Rate & $5 \times 10^{-5}$ & $5 \times 10^{-5}$ & $5 \times 10^{-5}$ \\
Batch Size (per GPU) & 2 & 1 & 1 \\
Num GPUs & 4 & 4 & 4 \\
Effective Batch Size & 8 & 4 & 4 \\
\midrule
\multicolumn{4}{l}{\textit{Generation Parameters}} \\
Generations per Prompt ($K$) & 8 & 8 & 5 \\
Temperature ($\tau$) & 1.0 & 1.0 & 1.0 \\
Max Prompt Length & 4096 & 4096 & 4096 \\
Max Completion Length & 2048 & 2048 & 2048 \\
\midrule
\multicolumn{4}{l}{\textit{Optimization Parameters (Shared)}} \\
Optimizer & \multicolumn{3}{c}{AdamW} \\
Gradient Clipping & \multicolumn{3}{c}{1.0} \\
LR Schedule & \multicolumn{3}{c}{Cosine with 10\% warmup} \\
Evaluation Frequency & \multicolumn{3}{c}{Every 50 steps} \\
\bottomrule
\end{tabular}
} 
\end{table}

\textbf{Training Algorithm.} We use Group Relative Policy Optimization (GRPO)~\citep{shao2024deepseekmath}, a batch-wise advantage estimation algorithm designed for mathematical reasoning. For each training example, we generate diverse completions and compute advantages by comparing rewards within each group. This approach reduces variance compared to single-sample methods while maintaining exploration. We optimize using AdamW with gradient clipping (max norm 1.0) and apply a cosine learning rate schedule with 10\% warmup.

\textbf{Reward Functions.}
We design task-specific reward functions that balance correctness, reasoning quality, and response format. We tried several formulations and report the ones that performed best so far, though they are not necessarily optimal and warrant further exploration.

\textbf{Counting Rewards.} Multi-component reward ($r \in [-0.4, +1.1]$) combining binary correctness ($r = 1.0$ correct, $r = 0.0$ incorrect), format quality bonuses ($+0.1$ for ``Answer: X'' format, $+0.05$ for acceptable variants, down to $-0.1$ for invalid format), and reasoning step penalties ($-0.1$ per step beyond 5 steps, capped at $-0.3$). These components apply to both correct and incorrect answers, creating positive rewards for well-formatted responses and negative rewards ($r \in [-0.4, 0]$) for verbose incorrect answers.

\textbf{Graph Reasoning Rewards.} Structured reward ($r \in [-0.2, +1.1]$) combining binary correctness ($r = 1.0$ correct, $r = 0.0$ incorrect) and format quality bonuses ($+0.1$ for proper \texttt{\{"answer": ...\}} JSON format). Incorrect but well-formatted responses receive partial credit ($r = 0.1$), while unstructured or excessively long outputs incur penalties ($r = -0.2$). This design reuses benchmark validation for consistent correctness evaluation.

\textbf{Spatial Reasoning Rewards.} Binary exact-match reward ($r \in \{0, 1\}$) using query-specific validation methods. Answer extraction supports flexible JSON formatting in multiple patterns (direct objects, code blocks) to accommodate diverse model response strategies.

\textbf{Training Configuration.}
Table~\ref{tab:rl_hyperparameters} summarizes our training hyperparameters. All models are trained on 4$\times$ NVIDIA A100 80GB GPUs, with training times ranging from 5--12 hours depending on problem complexity and dataset size.

\textbf{Evaluation Protocol.}
During training, we monitor validation performance every 50 steps on a held-out 10\% split from the training distribution. After training is complete, we evaluate on the curated test subset (Section~\ref{sec:curation}). We use greedy decoding (temperature 0) for test evaluation to assess the model's most confident predictions. Test accuracy serves as our primary metric for comparing scaling behaviors across dataset sizes and difficulty distributions.

\section{Results}

We performed a range of RL fine-tuning experiments with the curated datasets and training setup outlined in the previous section. Our goal is to empirically compare data curation strategies under fixed training constraints, rather than to isolate a single causal mechanism. In this section, we first break down the training results on a per-dataset basis, then discuss the high-level implications of fine-tuning on programmatically generated data. Aggregate test accuracies are reported in Table~\ref{tab:comprehensive_results}; per-difficulty breakdowns are provided alongside each dataset's training curves.

\subsection{Dataset Results}

\begin{figure*}[p]
  \centering

  \begin{subfigure}{\textwidth}
    \centering
    \begin{tabular}{cc}
      \includegraphics[width=0.49\textwidth]{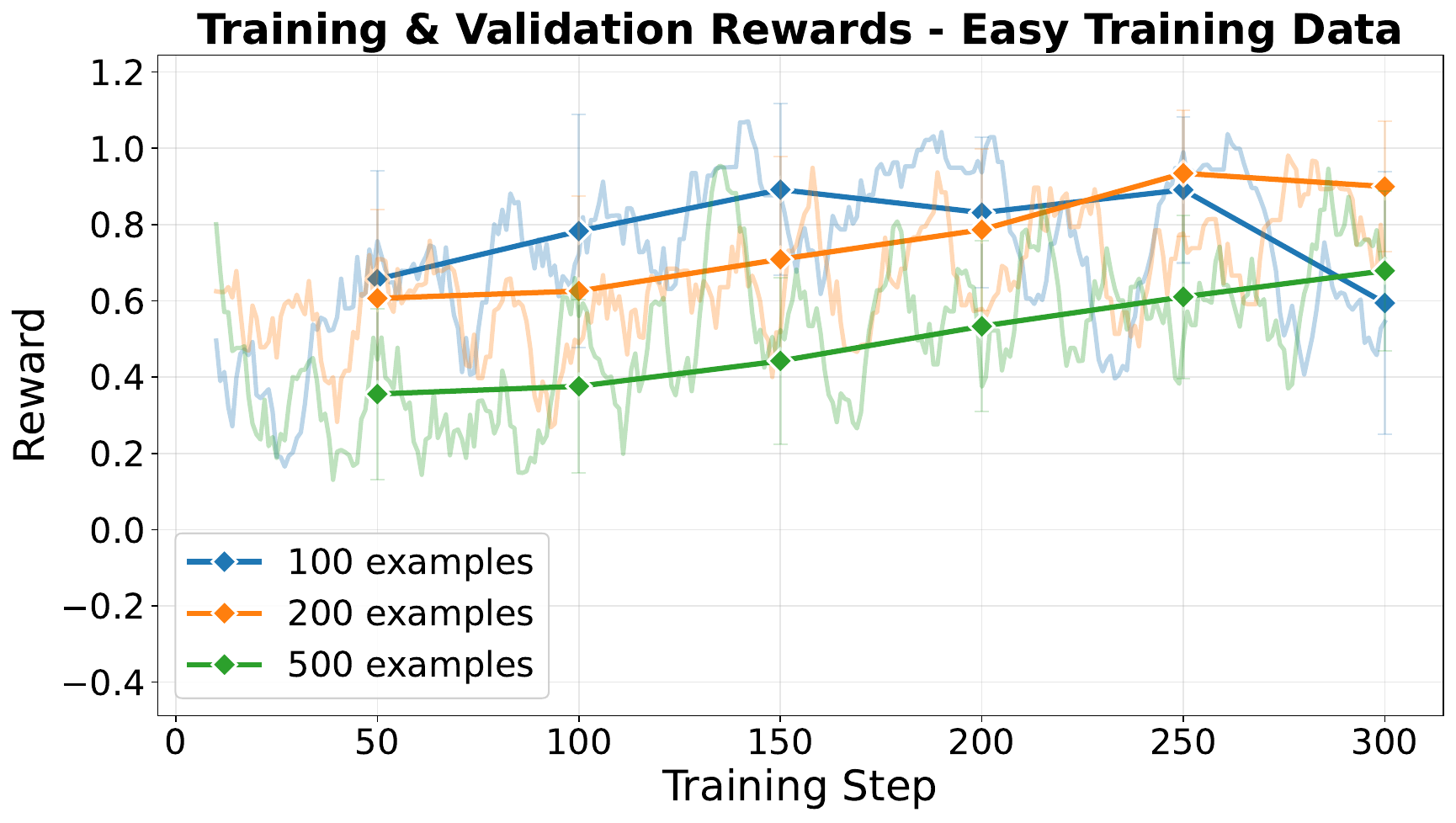} &
      \includegraphics[width=0.49\textwidth]{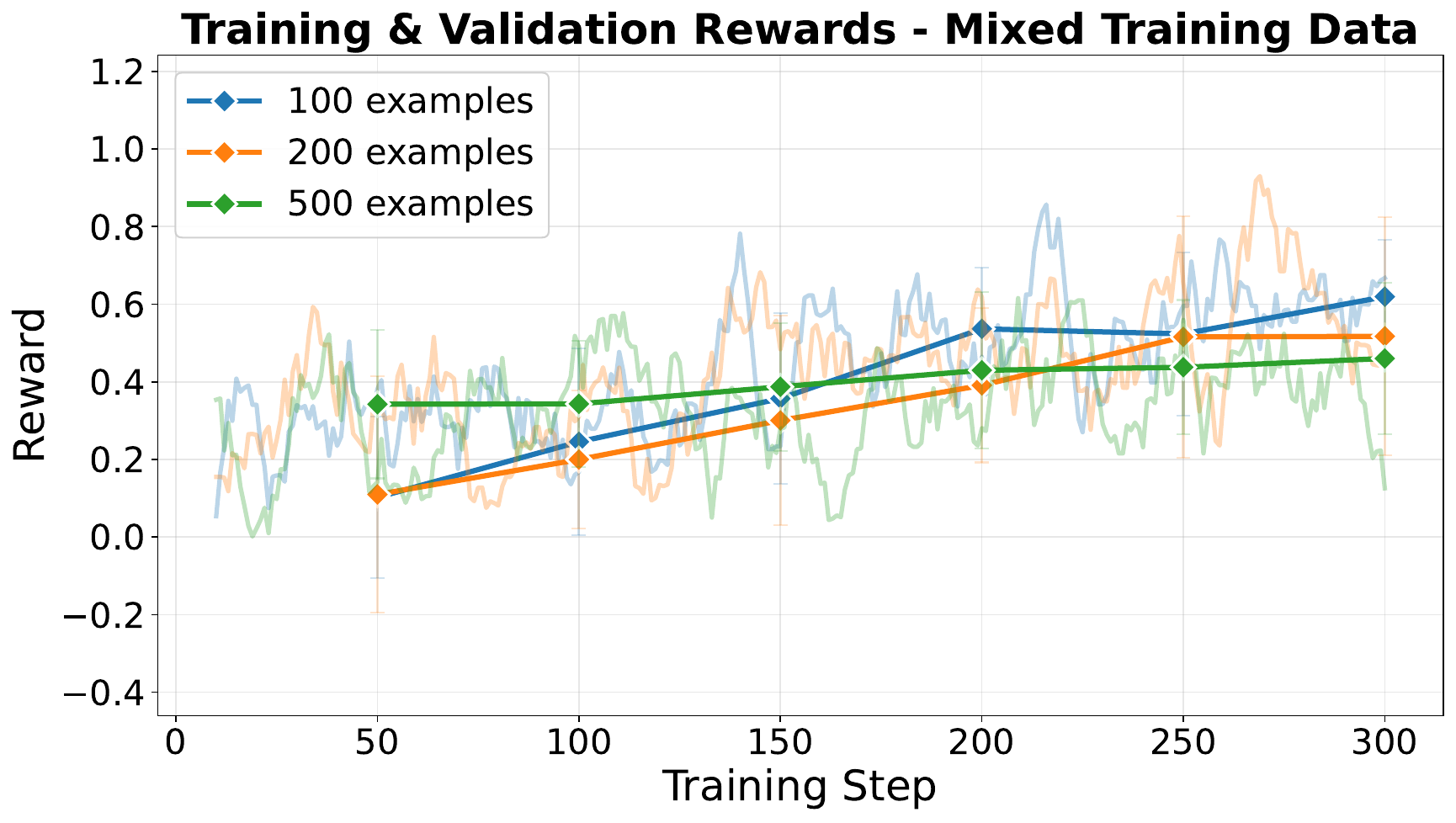} \\[-0.2em]
    \end{tabular}
    \caption{Counting: training and validation reward over 300 steps. Note the instability in the easy 100-example model (collapse after step 200).}
    \label{fig:counting_training}
  \end{subfigure}

  \vspace{0.3em}

  \begin{subfigure}{\textwidth}
    \centering
    \begin{tabular}{cc}
      \includegraphics[width=0.49\textwidth]{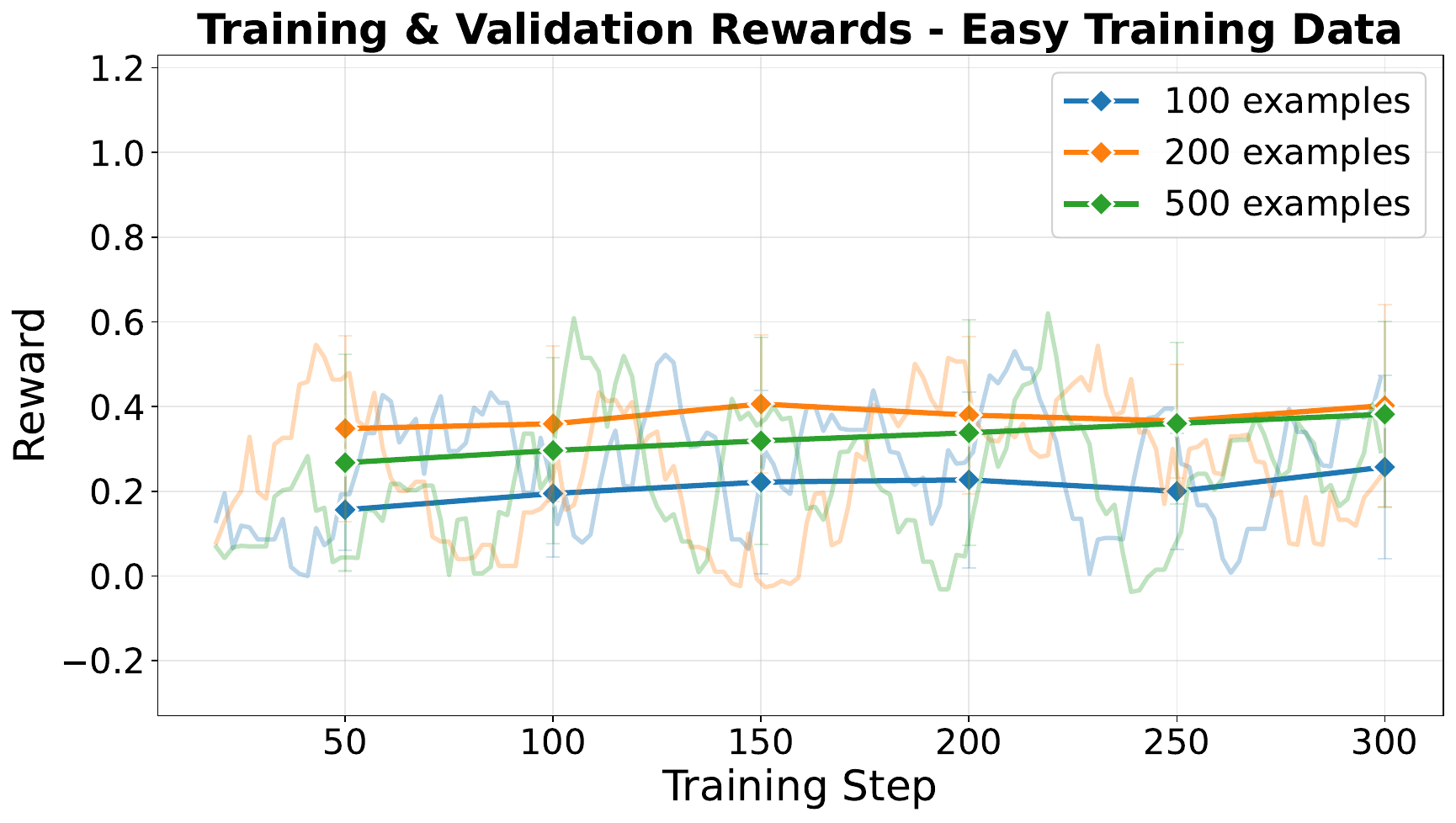} &
      \includegraphics[width=0.49\textwidth]{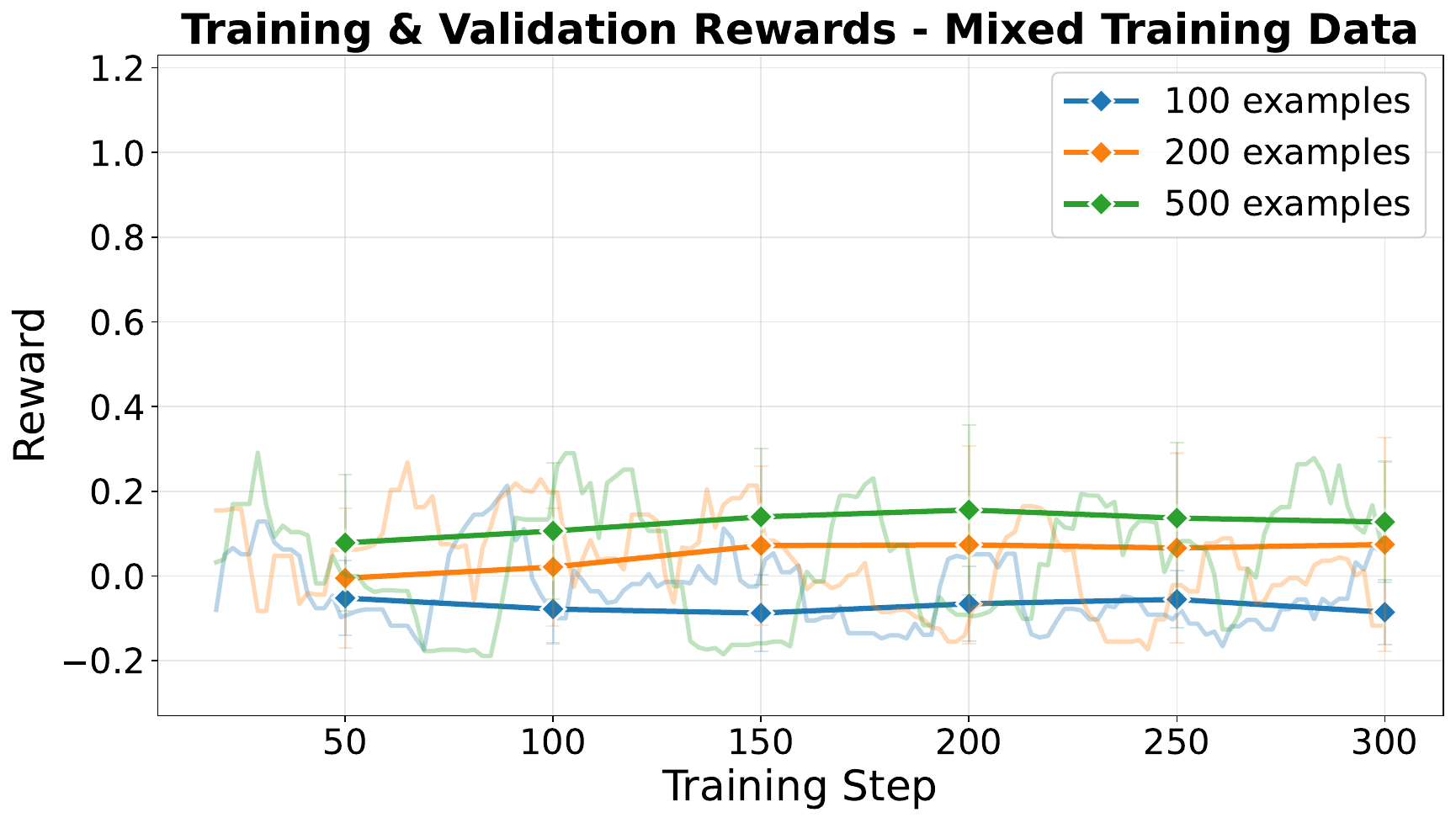} \\[-0.2em]
    \end{tabular}
    \caption{Graph: training and validation reward over 300 steps. Easy-only training (left) achieves positive validation rewards; mixed-difficulty (right) shows consistently negative rewards due to incomplete rollouts under token constraints.}
    \label{fig:graph_training}
  \end{subfigure}

  \vspace{0.3em}

  \begin{subfigure}{\textwidth}
    \centering
    \begin{tabular}{cc}
      \includegraphics[width=0.49\textwidth]{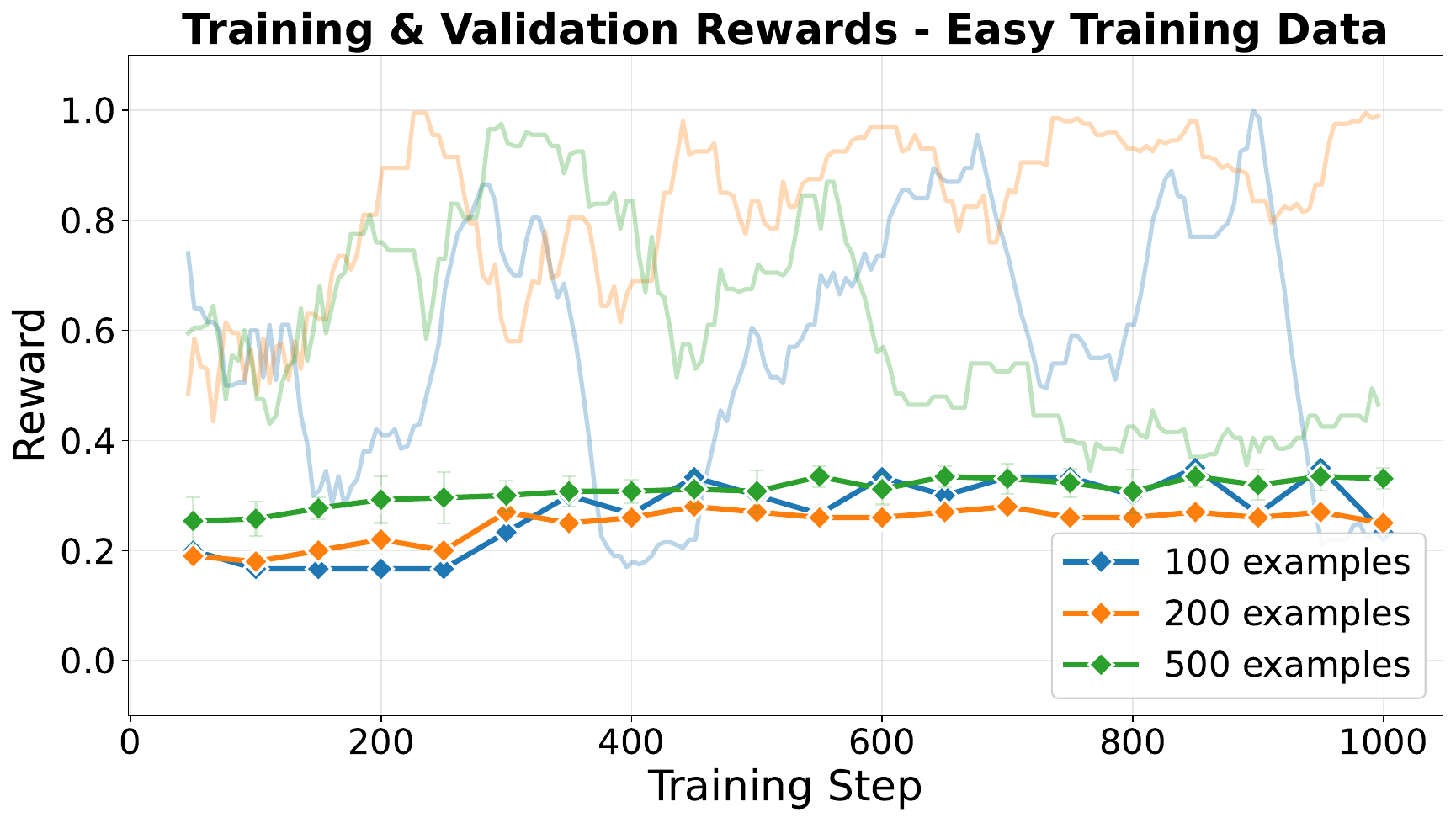} &
      \includegraphics[width=0.49\textwidth]{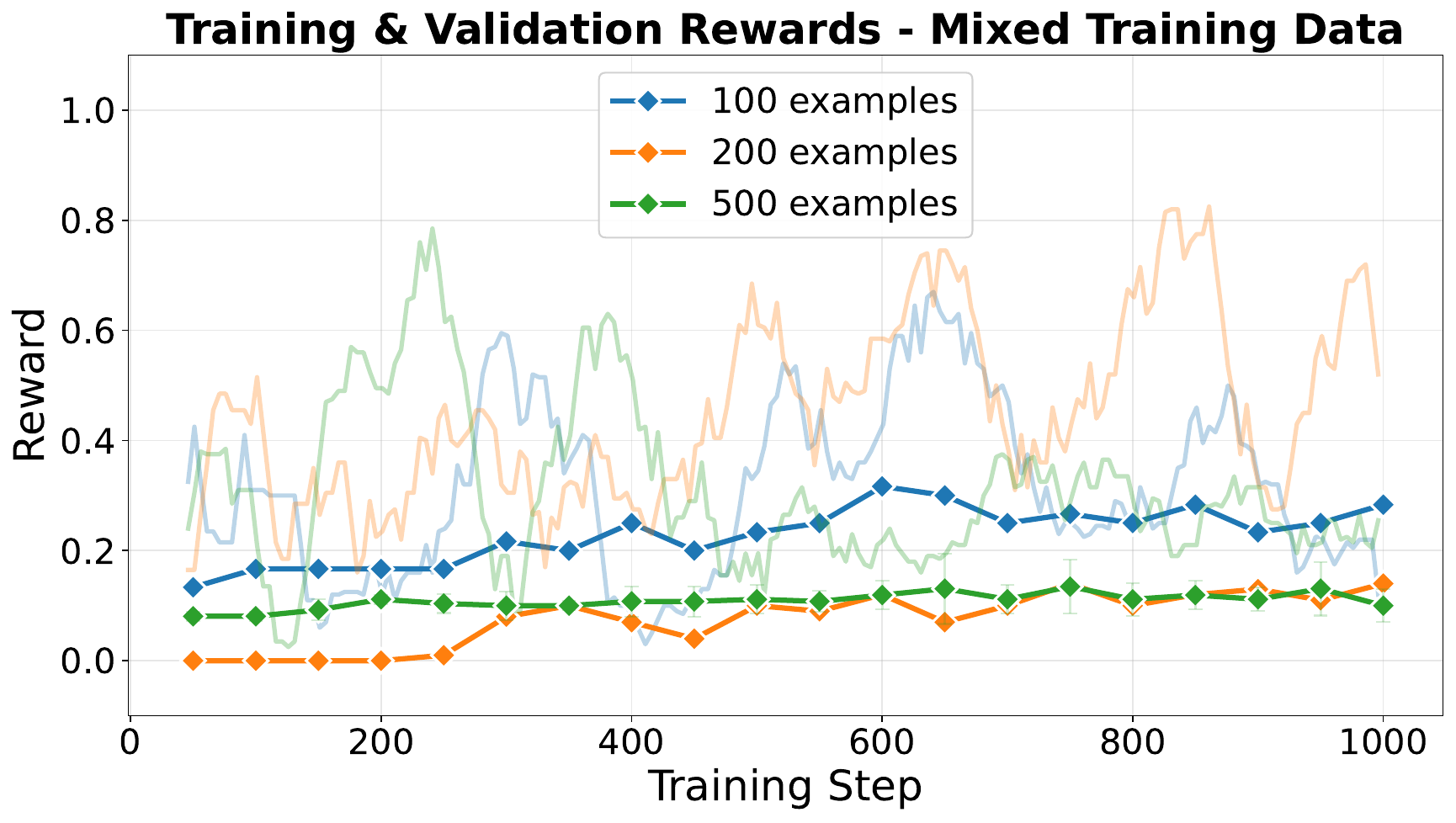} \\[-0.2em]
    \end{tabular}
    \caption{Spatial: training and validation reward over 1000 steps. Binary reward ($r \in \{0, 1\}$) creates discrete performance levels. Both regimes show steady improvement.}
    \label{fig:spatial_training}
  \end{subfigure}

  \caption{Training reward curves across all three datasets (left: easy-only, right: mixed-difficulty). Colors: blue = 100, orange = 200, green = 500 examples. Light shaded lines show training rewards; dark solid lines with diamond markers show validation rewards.}
  \label{fig:all_training}
\end{figure*}

\begin{figure*}[p]
  \centering

  \begin{subfigure}{\textwidth}
    \centering
    \begin{tabular}{cc}
      \includegraphics[width=0.46\textwidth]{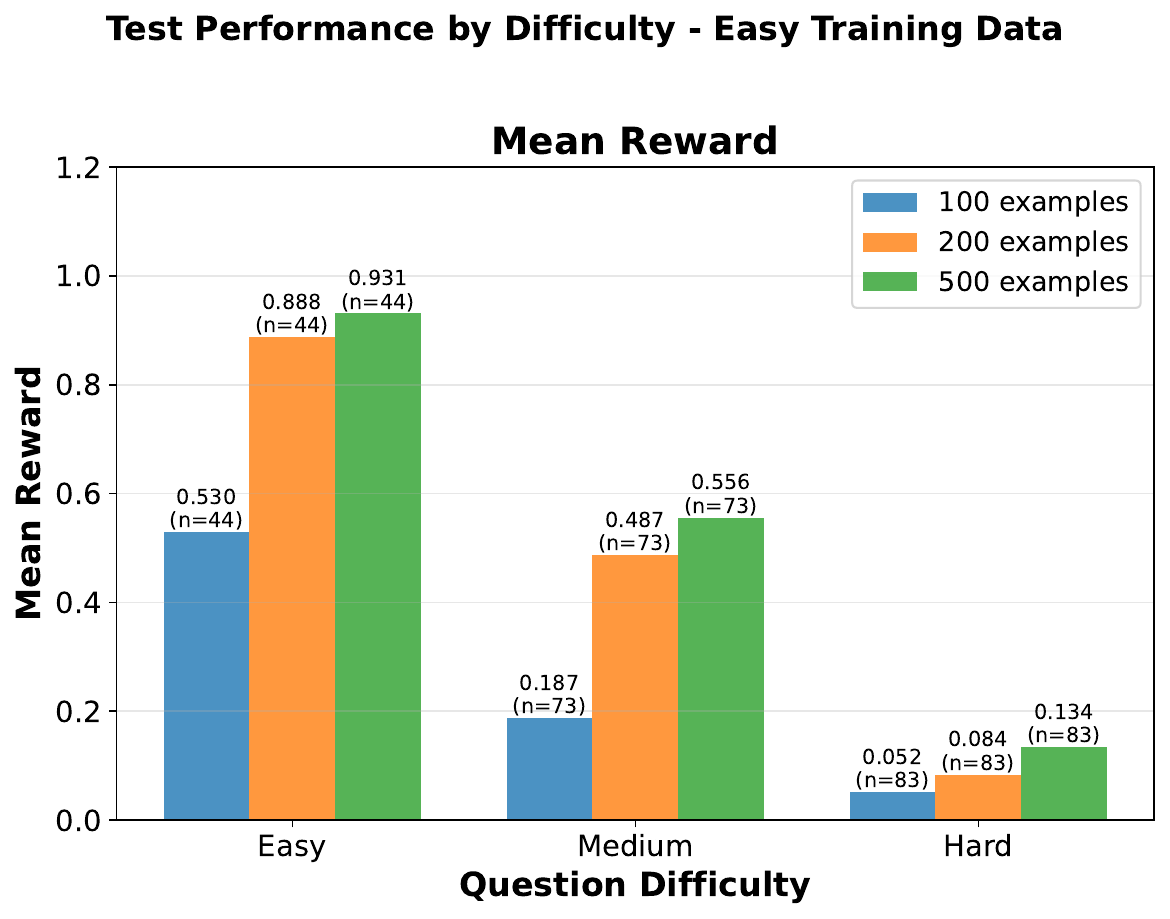} &
      \includegraphics[width=0.46\textwidth]{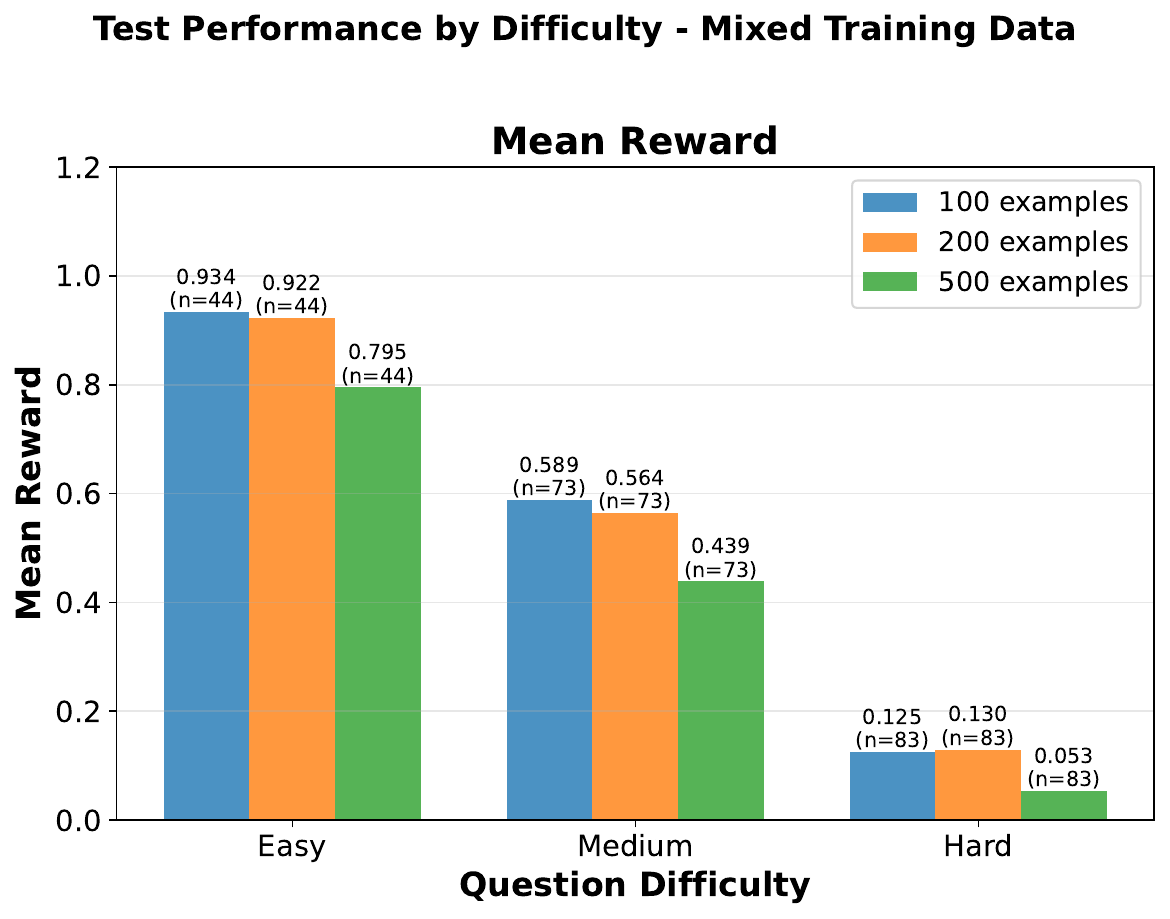} \\
    \end{tabular}
    \caption{Counting: easy-trained models (left) specialize on easy questions but fail on harder ones; mixed-trained models (right) maintain consistent cross-difficulty performance.}
    \label{fig:counting_test}
  \end{subfigure}

  \vspace{0.3em}

  \begin{subfigure}{\textwidth}
    \centering
    \begin{tabular}{cc}
      \includegraphics[width=0.46\textwidth]{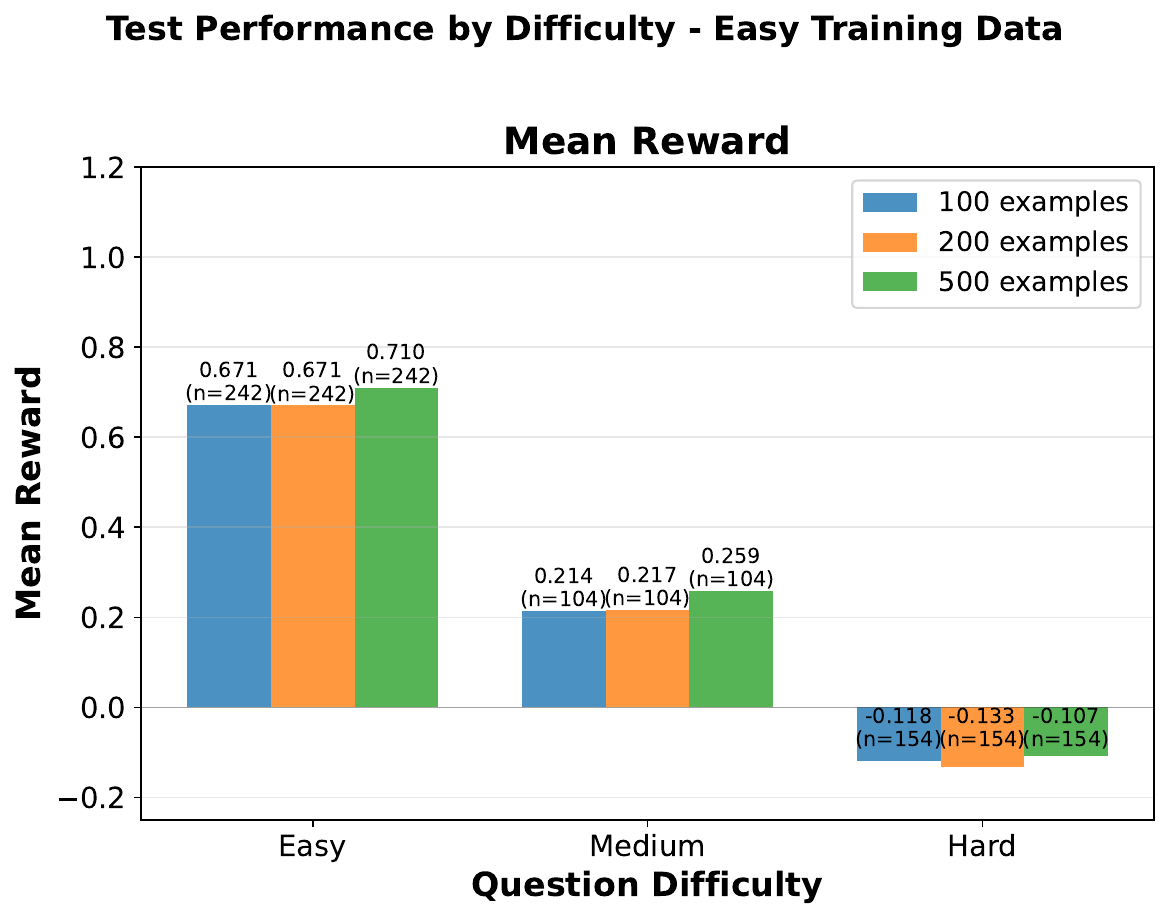} &
      \includegraphics[width=0.46\textwidth]{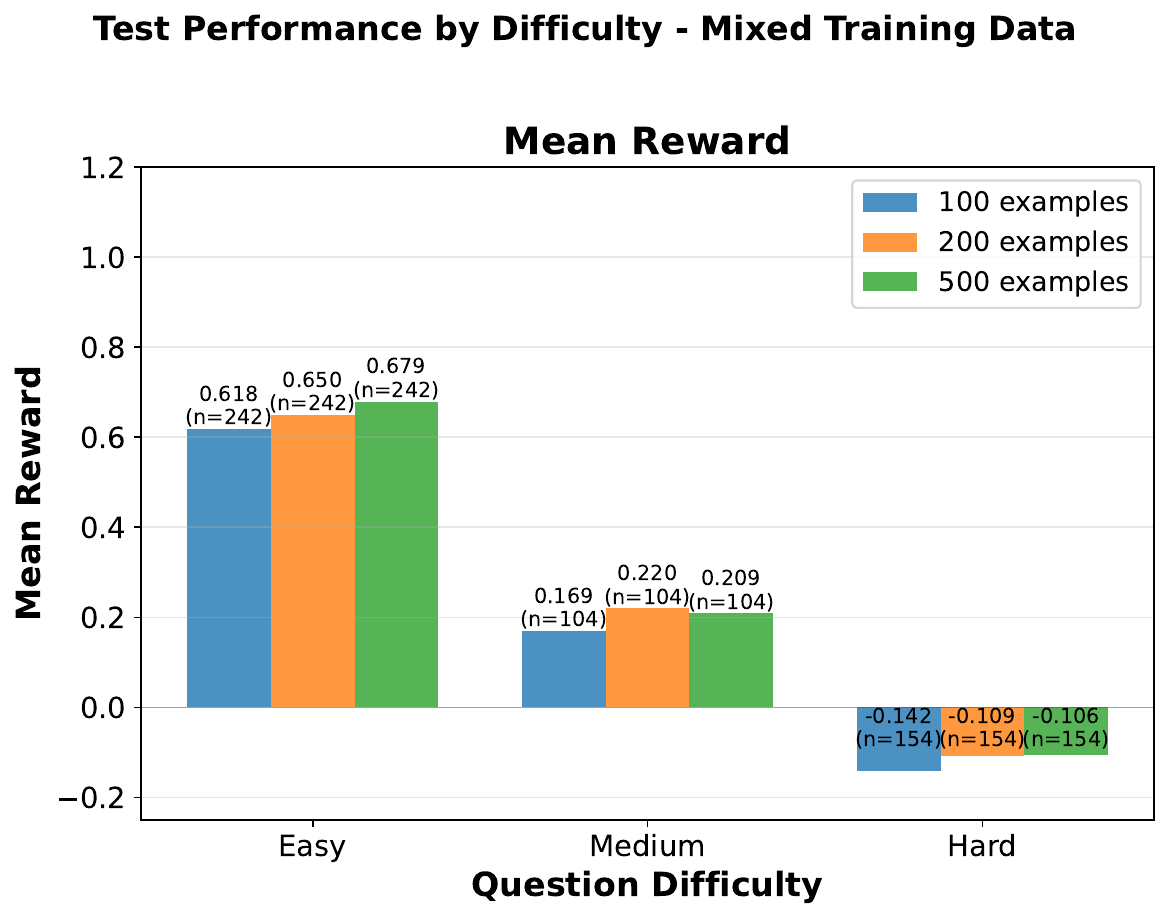} \\
    \end{tabular}
    \caption{Graph: both training regimes achieve positive rewards on easy questions but struggle with medium and hard problems due to token constraints.}
    \label{fig:graph_test}
  \end{subfigure}

  \vspace{0.3em}

  \begin{subfigure}{\textwidth}
    \centering
    \begin{tabular}{cc}
      \includegraphics[width=0.46\textwidth]{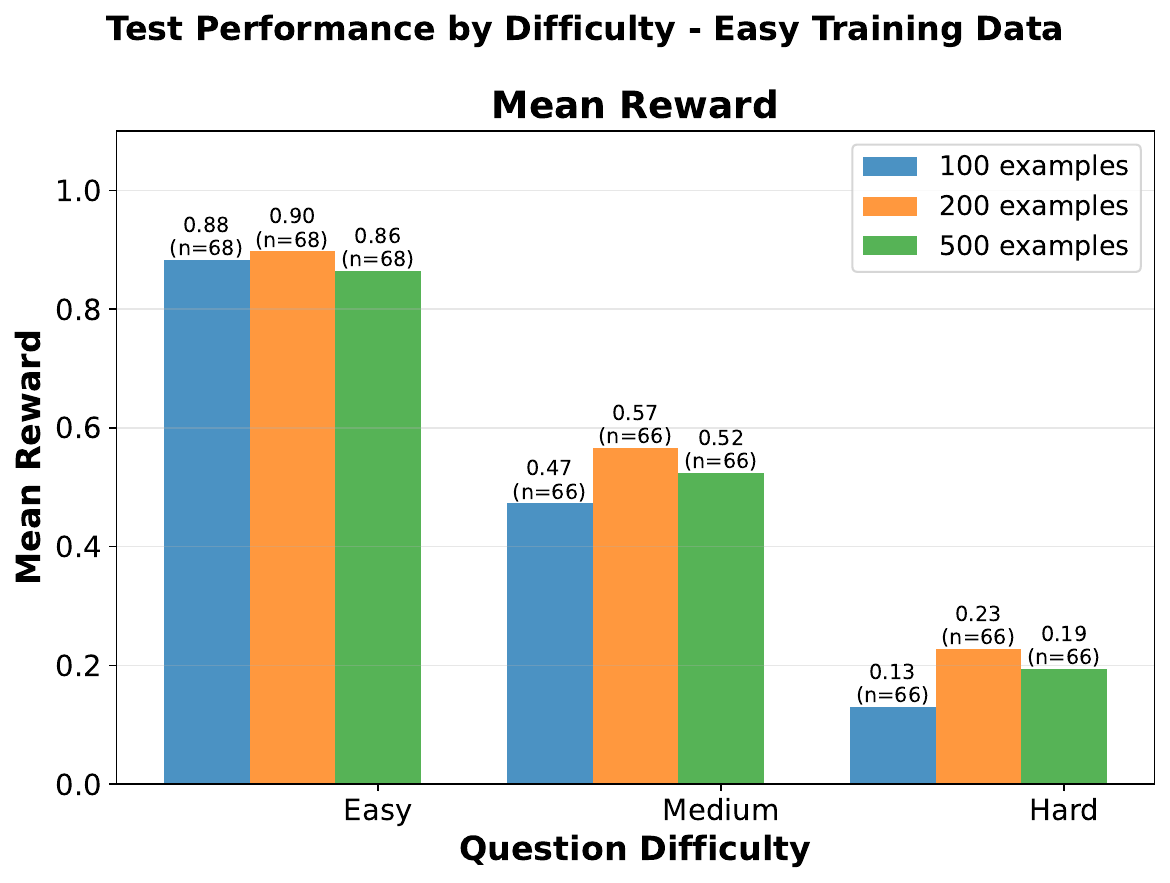} &
      \includegraphics[width=0.46\textwidth]{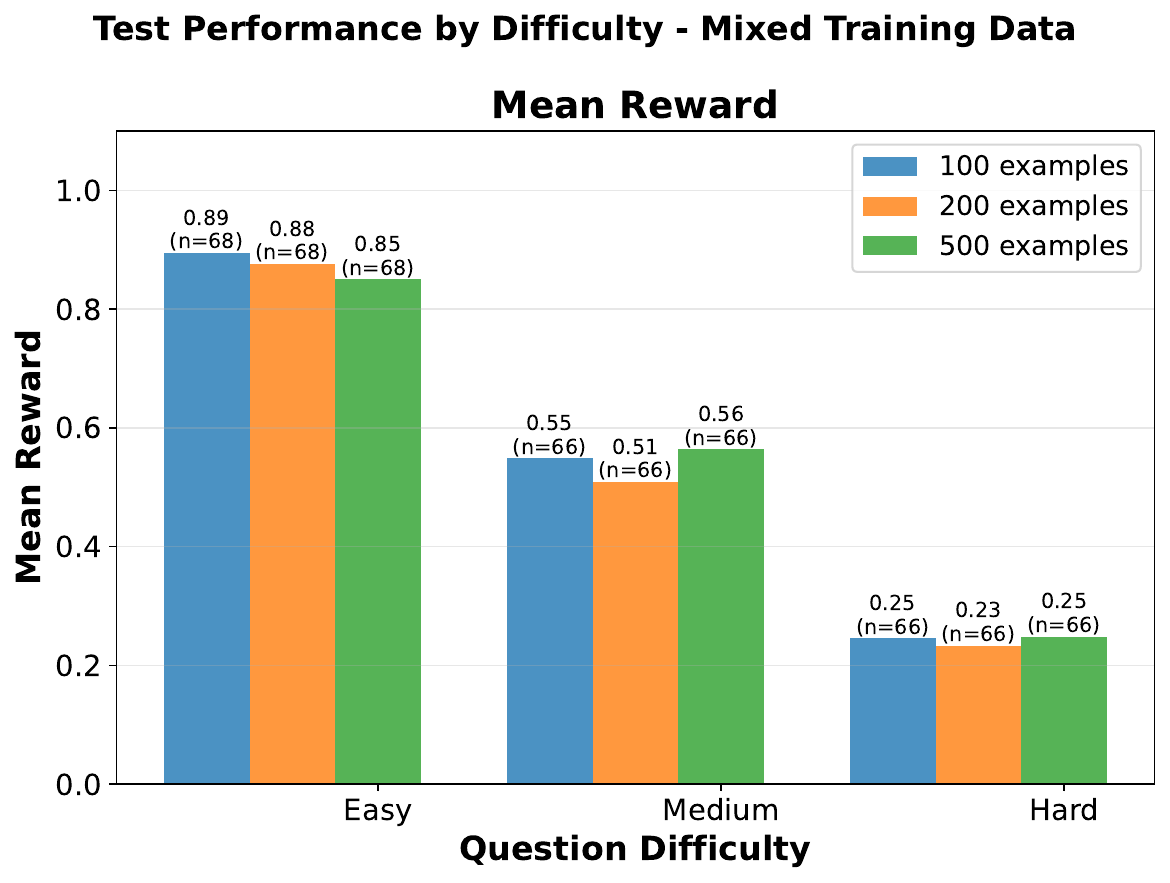} \\
    \end{tabular}
    \caption{Spatial: discrete binary rewards yield modest cross-distribution differences, with both models generalizing similarly.}
    \label{fig:spatial_test}
  \end{subfigure}

  \caption{Test accuracy by question difficulty across all three datasets (left: easy-trained, right: mixed-trained). Colors: blue = 100, orange = 200, green = 500 examples. Bars show accuracy on the held-out test set.}
  \label{fig:all_test}
\end{figure*}

\subsubsection{Counting Problems}

\textbf{Training-Time Observations.}
Figure~\ref{fig:counting_training} shows the training progression for both easy-difficulty and mixed-difficulty. Under a fixed 300-step training budget, models trained on mixed-difficulty data overall showed continued validation improvement through step 300, suggesting that larger datasets may require proportionally more training steps---under compute constraints, small diverse datasets outperform large homogeneous ones.

\begin{figure}[b]
  \centering
  \includegraphics[width=\linewidth]{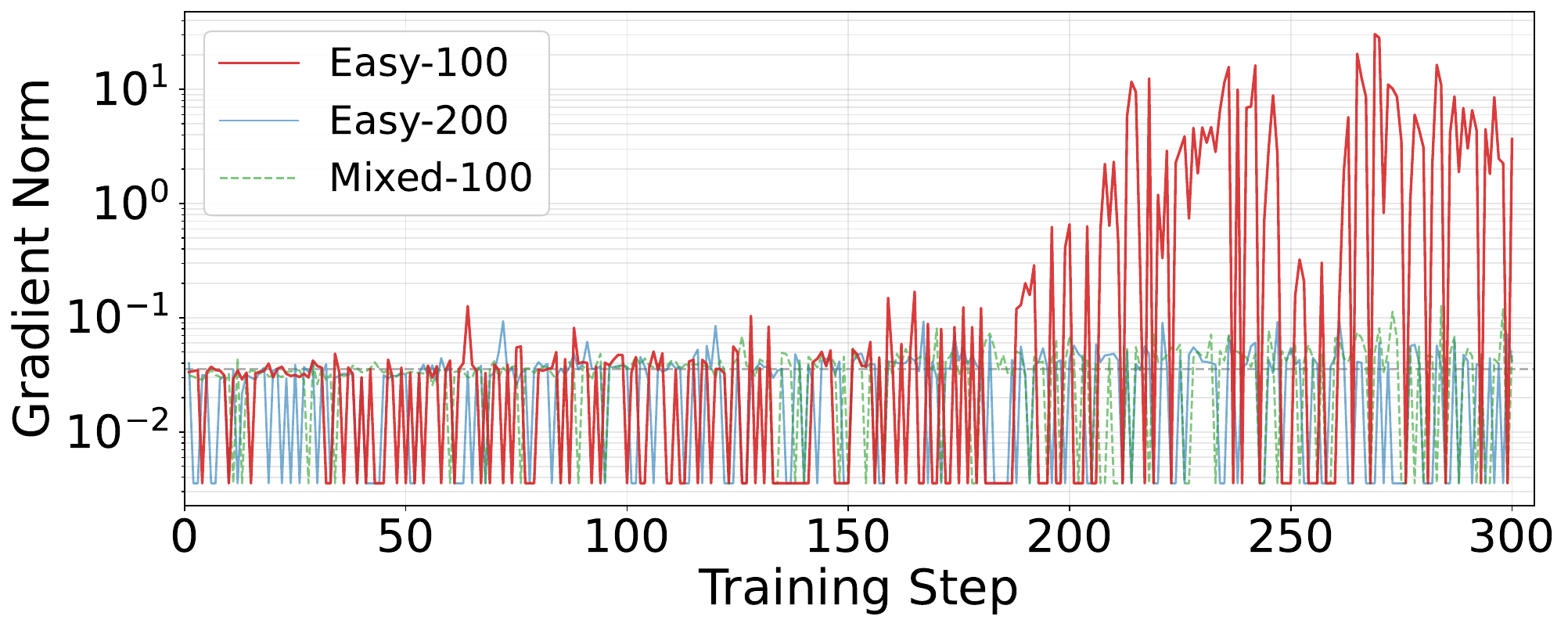}
  \caption{Gradient norm over training steps for three counting configurations. Easy-100 exhibits spikes exceeding 850$\times$ baseline between steps 150--300, coinciding with the reward collapse in Figure~\ref{fig:counting_training}. Easy-200 and Mixed-100 remain stable throughout, supporting a minimum diversity threshold for stable optimization.}
  \label{fig:counting_gradnorm}
\end{figure}

\begin{figure}[t]
  \centering
  {\small\textbf{Counting Problems}}\par\vspace{0.2em}
  \includegraphics[width=\linewidth]{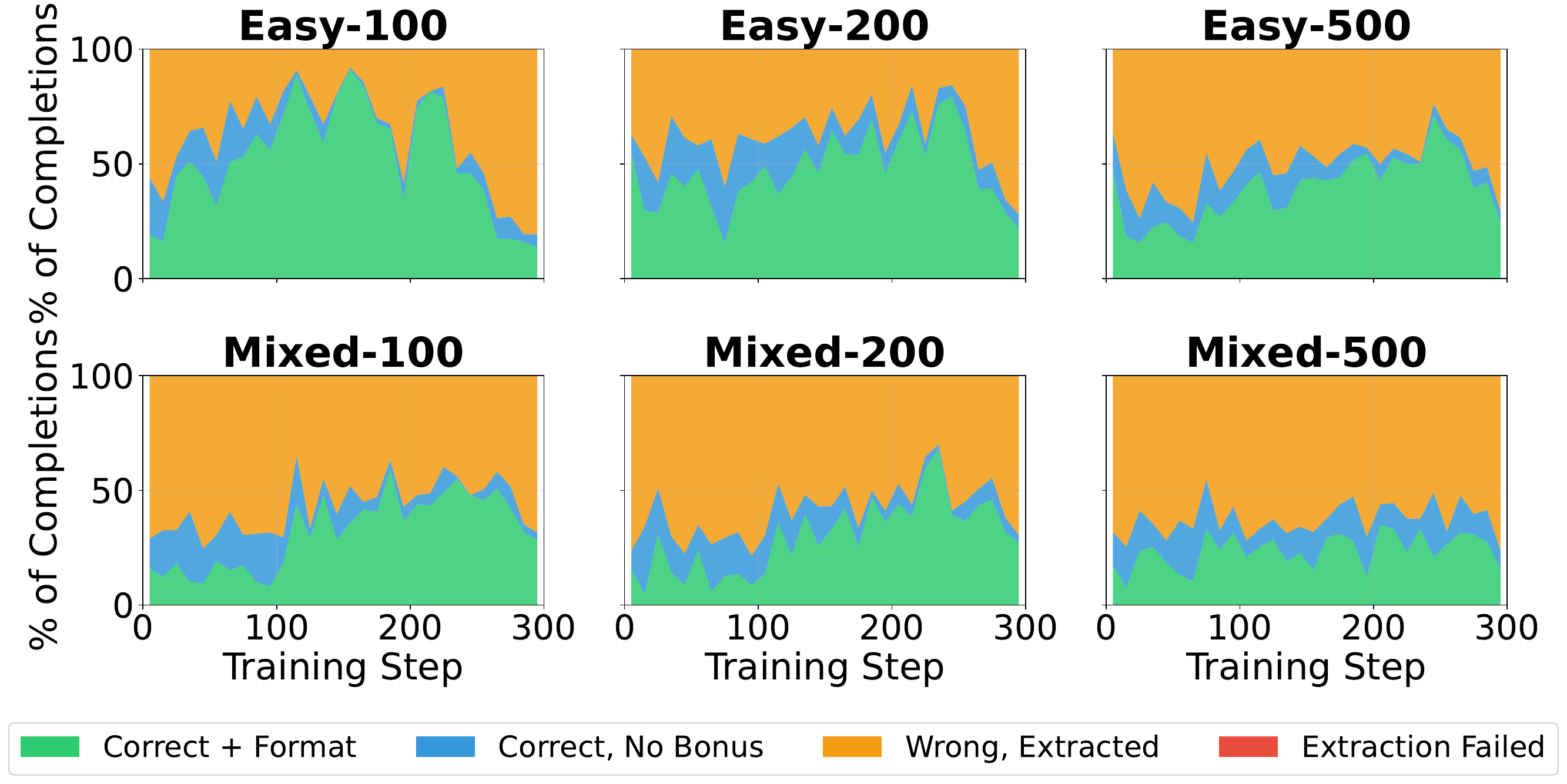}

  \vspace{0.5em}

  {\small\textbf{Graph Reasoning}}\par\vspace{0.2em}
  \includegraphics[width=\linewidth]{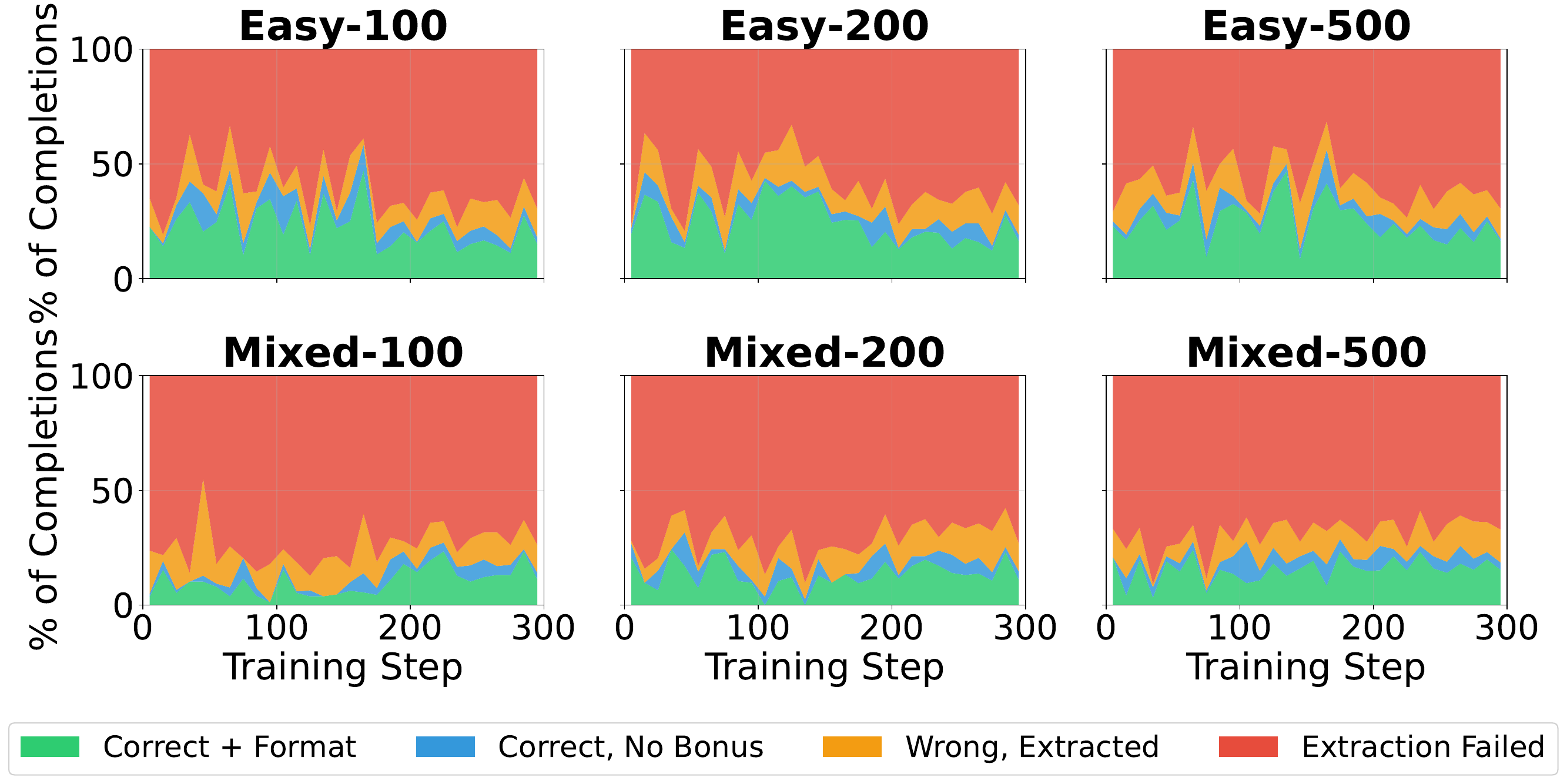}
  \caption{Reward component breakdown across training configurations. For counting, Easy-100 correctness collapses after step 150, consistent with gradient norm instability (Figure~\ref{fig:counting_gradnorm}). For graph reasoning, extraction failures dominate (59--65\% easy, 67--73\% mixed), explaining persistently negative aggregate rewards. Spatial reasoning is excluded (binary reward, no sub-components).}
  \label{fig:reward_components}
\end{figure}

Easy-difficulty training showed greater variation across dataset sizes. The 100-example model exhibited severe instability, peaking at 0.89 validation reward (step 150) before declining to 0.59 (step 300), coinciding with gradient norm spikes exceeding 850$\times$ baseline values (Figure~\ref{fig:counting_gradnorm}). The 200- and 500-example models trained stably (final rewards 0.90 and 0.68), suggesting a minimum dataset size threshold between 100 and 200 examples for stable optimization. Notably, Mixed-100 trained stably despite the same sample count, indicating that difficulty diversity can substitute for dataset size in stabilizing RLVR training under limited data and compute. To better understand what drives aggregate reward variation, we decompose completions into correctness and format categories (Figure~\ref{fig:reward_components}). For counting, the breakdown confirms that the Easy-100 collapse is driven by a drop in correctness rate rather than format or extraction issues.

\textbf{Test-Time Generalization.}
Test performance confirmed consistent improvement for the easy models, and showed a degradation in performance for mixed models. Mixed models achieved 0.478 (100 examples), 0.476 (200 examples), and 0.367 (500 examples) mean reward, with corresponding solving accuracies of 50.0\%, 50.5\%, and 40.0\%. Easy-trained models scaled monotonically from 0.218 to 0.461, but required 5$\times$ more examples to match mixed baselines (500 easy examples $\approx$ 100 mixed examples in final accuracy).

Figure~\ref{fig:counting_test} decomposes test performance by question difficulty. Mixed-trained models show performance degradation with scaling. In contrast, easy-trained models scale monotonically. The 100-example mixed model maintains the most balanced cross-difficulty profile, while easy-trained models require 500 examples to match mixed-trained performance on easy questions.

Results demonstrate two novel scaling behaviors: (1) \textit{Degraded scaling under compute constraints}---however, with continued validation improvement toward the final step, we hypothesize that with more compute, we could see potential inverted-U scaling, contradicting supervised fine-tuning scaling laws that predict monotonic improvement~\citep{zhang2024scaling}; and (2) \textit{5$\times$ sample efficiency of diverse training data}, suggesting that data composition may outweigh data quantity in low-resource RL regimes. These findings suggest that practitioners facing compute constraints should prioritize dataset diversity over size.
\begin{table}[h!]
\centering
\small
\scalebox{0.92}{
\begin{tabular}{ccccccc}
\toprule
\multirow{2}{*}{\textbf{Samples}} & \multicolumn{2}{c}{\textbf{Counting}} & \multicolumn{2}{c}{\textbf{Graph}} & \multicolumn{2}{c}{\textbf{Spatial}} \\
\cmidrule(lr){2-3} \cmidrule(lr){4-5} \cmidrule(lr){6-7}
 & \textbf{Easy} & \textbf{Mixed} & \textbf{Easy} & \textbf{Mixed} & \textbf{Easy} & \textbf{Mixed} \\
\midrule
0 & 31.3 & 31.3 & 29.4 & 29.4 & 26.1 & 26.1 \\
\cmidrule(lr){1-7}
100 & 21.9 & 44.2 & 33.3 & 29.1 & 49.9 & 56.6 \\
\cmidrule(lr){1-7}
200 & 40.0 & 43.4 & 32.9 & 32.7 & 56.7 & 54.3 \\
\cmidrule(lr){1-7}
500 & 44.2 & 35.5 & 36.5 & 34.0 & 53.1 & 55.7 \\
\bottomrule
\end{tabular}
}
\caption{Mean test accuracy (\%) vs training sample size across all datasets and difficulty settings. First row shows base model performance (Qwen3-4B with no fine-tuning).}
\label{tab:comprehensive_results}
\end{table}

\subsubsection{Graph Reasoning}

\textbf{Training-Time Observations.}
Graph reasoning showed stable but modest improvement in validation performance across all training runs, except for the \textit{Mixed-100} configuration (Figure~\ref{fig:graph_training}). In contrast, all easy-only runs showed mostly consistent upward reward trajectories through step 300, indicating continued learning despite high training reward variance. As shown in Table~\ref{tab:comprehensive_results}, the \textit{Easy-500} model achieved the strongest test performance overall, suggesting that reinforcement learning was more stable and effective on larger, easier datasets in this setting.

Mixed-difficulty training was more challenging. Validation rewards were frequently negative across all dataset scales (Figure~\ref{fig:graph_training}), reflecting incomplete rollouts where the model exhausted its token budget before producing a parseable structured output. On average, mixed datasets contained larger graphs (14.9 nodes vs. 12.6 for easy sets), resulting in longer input and output sequences that more often exceeded the maximum generation length. Since the reward function penalized cutoffs (Section~\ref{sec:reinforce}), these incomplete rollouts suppressed training rewards and slowed convergence. The reward component breakdown (Figure~\ref{fig:reward_components}) confirms that extraction failures account for the majority of completions across all graph configurations (59--65\% for easy, 67--73\% for mixed), explaining the persistently negative aggregate rewards. Notably, the \textit{Mixed-100} model performed slightly worse than the baseline (29.1\% vs.\ 29.4\%), coinciding with limited diversity and frequent negative reward signals from incomplete rollouts.

\textbf{Test-Time Generalization.}
Easy-trained models slightly outperformed mixed-trained models overall at test time (Figure~\ref{fig:graph_test}). However, both training setups failed to generalize beyond the easy test subset. Performance declined sharply on medium problems and became mostly negative on hard problems, indicating that the models struggled to reason over longer or more complex graphs. This is consistent with training inefficiencies under compute and context constraints, which limited the model's ability to complete reasoning and reduced opportunities to receive positive reward signals on more difficult rollouts.

Scaling analysis across 100--500 examples revealed only incremental improvements within each training regime (Figure~\ref{fig:graph_training}). Easy-trained models scaled roughly linearly in final validation reward, while mixed-trained models showed minimal change. This weak scaling trend supports the conclusion that \textit{training on harder instances---which involve longer sequences and more reasoning steps---is constrained by token limits and fixed compute budgets}.

Overall, graph reasoning scaling was constrained more by token generation limits than by data volume, given the verbose nature of the problems. Larger graphs inflated input and output lengths, causing the model to more frequently exhaust its token budget before producing complete outputs. Increasing these limits would change the underlying training dynamics and may yield different scaling behavior, making length-adaptive optimization an important direction for future work.

\subsubsection{Spatial Reasoning}

\begin{figure}
    \centering
    \includegraphics[width=\linewidth]{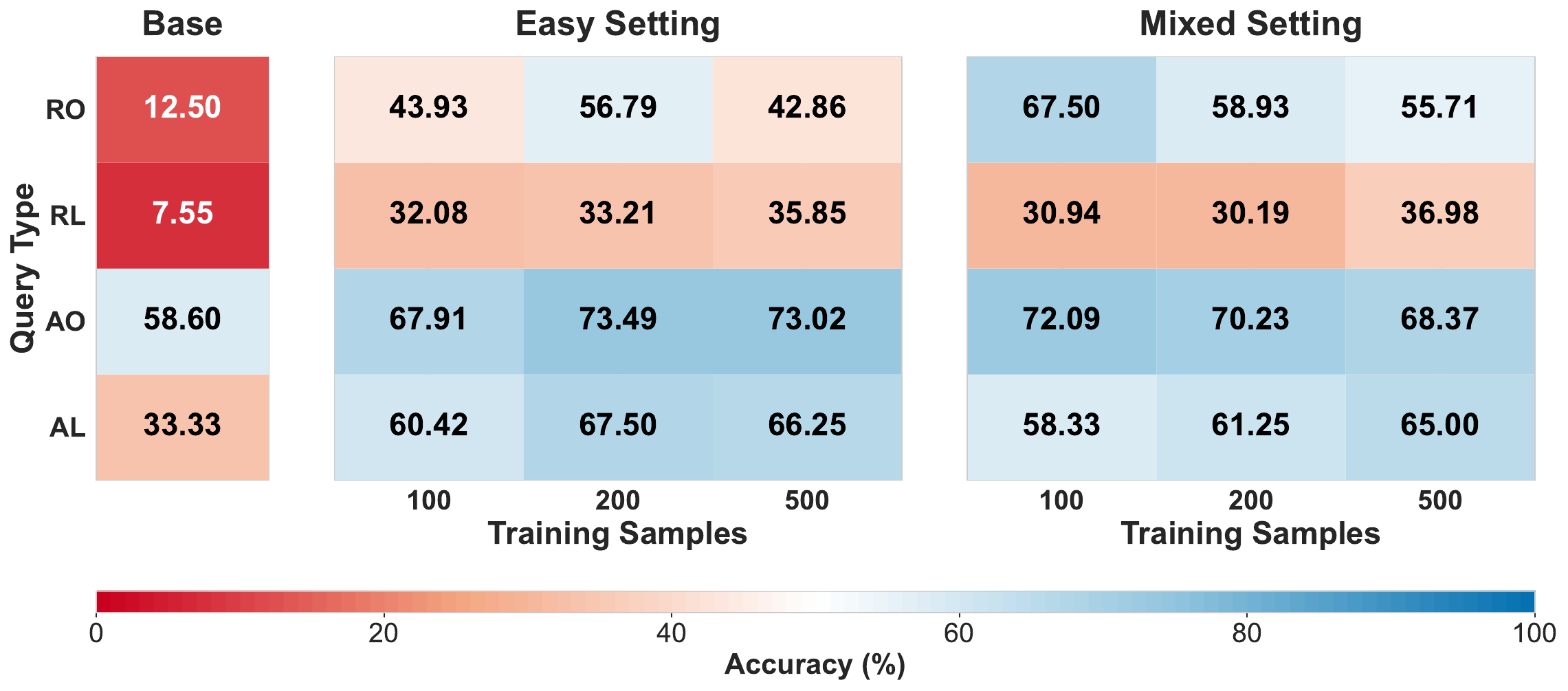}
    \caption{Test accuracy on different types of queries in spatial reasoning. Here, AO, AL stand for absolute orientation and absolute location, and similarly, RO, RL stand for relative queries. In both easy and mixed settings, we see improvements across all types of queries. The improvements are more pronounced on location and relative orientation queries. Moreover, in the mixed setting, the performance is much better on the relative orientation queries.}
    \label{fig:query_type_heatmap}
\end{figure}

\textbf{Training-Time Observations.} During training, we see that the validation accuracy improves gradually in all settings (Figure~\ref{fig:spatial_training}). The rate of improvement varies across the settings---the model learns faster in easy settings with fewer samples in comparison to other settings with more samples, especially with medium and hard difficulty levels. This is expected behavior with a fixed training budget across these settings; medium and hard questions are expected to be harder to learn and may require a longer training time and larger token limits.

\textbf{Test-Time Generalization.}
After fine-tuning the model, we evaluate it on a test set of 200 problems of mixed difficulty levels (68 easy, 66 medium, and 66 hard) spanning 4 query types: absolute location (48 questions), absolute orientation (43), relative location (53), and relative orientation (56). We draw the following insights from the results on the test data.

\textit{Fine-tuning with either type of training set improves accuracy.} While this is expected from fine-tuning in general, as a first step, it is necessary to study whether fine-tuning is effective in this novel setting of spatial reasoning. These results (in Table~\ref{tab:comprehensive_results}) provide clear evidence that fine-tuning improves the model's performance (up to 2$\times$) on the given spatial reasoning task.

\textit{Diminishing returns with more training data under compute constraints.}
Contrary to conventional wisdom, more training data may not yield higher accuracy when the compute budget for training is fixed. Our results on spatial reasoning in both easy and mixed training settings in Table~\ref{tab:comprehensive_results} support this claim. We see that in the easy setting the accuracy improves by about 7\% when moving from 100 to 200 training samples, but drops by 3.6\% in the 500 samples case. In the mixed setting, the accuracy does not change much with the number of samples, but we see a small dip when compared to the 100 samples setting. These results suggest an inverted U-shape scaling of test accuracy with the training set size under compute constraints.

\textit{Training on mixed difficulty samples generally performs better in comparison to training only on easy samples.}
 In Table~\ref{tab:comprehensive_results}, we observe that training on samples with mixed difficulty results in a higher or comparable test accuracy, in contrast to training on an easy set of the same size. Moreover, the accuracy with 100 mixed samples even surpasses the accuracy with 500 easy samples, suggesting training on fewer samples of varying difficulty is preferable to training on a large set of easy samples.

\textit{Improvements across all types of queries.}
Next, we study the performance of the model on different types of queries before and after fine-tuning. The results are shown in Figure~\ref{fig:query_type_heatmap}. We observe that fine-tuning with either training set improves accuracy across all query types. The baseline results suggest that relative queries are harder, and it is interesting to see that even training on easy samples increases the accuracy on these queries significantly. Similar improvements are also noted in the mixed setting, which are more pronounced on the relative orientation queries. This may reflect the fact that the mixed set contains more such queries in comparison to the easy set.

\textbf{Robustness considerations.} Due to computational cost, we did not perform multi-seed repetitions. We instead rely on consistency of qualitative trends across 18 configurations (3 domains $\times$ 6 data configurations) as a robustness signal, suggesting the observed effects are not solely driven by training noise. The mixed-vs-easy sample efficiency trend emerges independently in both Counting and Spatial Reasoning despite differing reward structures, and all configurations share fixed training budgets, token limits, and task-specific reward functions, so performance differences reflect how dataset size and composition interact with these constraints. We view these findings as empirical observations under low data and compute regimes that generate hypotheses rather than definitive causal claims.

\subsection{Implications}

Our results across the datasets point to a unifying theme: under fixed-budget RLVR regimes, performance is shaped by interactions between dataset size, composition, training duration, and token limits. We distill three design lessons for practitioners.

\textbf{1. Training-set composition can outweigh data volume.}
In both Counting and Spatial Reasoning, small mixed-difficulty datasets matched or exceeded the test accuracy of larger easy-only datasets. For Counting, 100 mixed examples matched 500 easy examples in test accuracy, a \textbf{5$\times$ sample efficiency advantage}, suggesting that curating across empirically defined difficulty tiers can be more effective than scaling up easy examples alone.

\textbf{2. Under fixed training budgets, scaling data alone can be ineffective.}
Larger datasets receive fewer optimization updates per example under a fixed step budget. In Counting, mixed-difficulty test accuracy declined beyond 100 examples even though validation rewards were still improving at the final step. In Spatial Reasoning, easy-trained accuracy peaked at 200 examples before declining, and mixed-trained accuracy did not improve beyond 100 examples. These patterns motivate joint consideration of dataset size, training duration, and token budgets.

\textbf{3. Harder instances can be constrained by incomplete rollouts under fixed token limits.}
In Graph Reasoning, larger graphs frequently exhausted the token budget before generating parseable outputs, suppressing positive reward signals (Figure~\ref{fig:reward_components}). Counting and Spatial Reasoning, with shorter outputs, were less affected. For domains requiring verbose reasoning, token budget allocation may be a more binding constraint than data size.

Consistent with the multi-factor framing in Section~\ref{sec:datasets}, these findings reflect interacting system constraints rather than effects of a single ``complexity'' factor. Alternative contributors, including optimization dynamics, reward sparsity, and task-specific reward design, may also shape the observed patterns.

\section{Conclusion}
In this work, we have presented a systematic characterization of open-source SLMs fine-tuned using RLVR under low data regimes. Using three procedurally generated datasets with controllable properties (covering number counting problems, graph reasoning, and spatial reasoning), we characterize the effects of dataset size, diversity, and complexity on model reasoning capabilities. We find that even within low data regimes, under RLVR, open-source SLMs trained on low complexity tasks generalize to higher complexity tasks not seen during training, and training on a mixture of complexities is associated with greater accuracy gains under lower data budgets---specifically, we find that the mixed complexity setting provides up to $5\times$ the sample efficiency versus training on easy tasks alone. Our results also demonstrate that procedural data generation is a useful tool for starting to understand the data scaling laws that govern the effectiveness of RLVR, and these findings motivate wider research into the development of empirical laws that relate reasoning capabilities after fine-tuning to dataset composition and size.

Although our work begins to characterize the effectiveness of RLVR in low-data regimes, several limitations should be noted. Our findings are specific to a 4B parameter model with LoRA fine-tuning under fixed low compute budgets, and we did not perform multi-seed repetitions. These results are intended to inform practitioners operating under similar constraints, not to claim universal scaling laws. While data composition effects may remain relevant at larger scales~\citep{tan2025scaling}, the quantitative gains (e.g., 5$\times$ sample efficiency) may not directly transfer to larger models. Additionally, procedural tasks do not capture the full complexity of real-world data, and we did not evaluate transfer to natural language benchmarks.

Future work could develop more extensive scaling laws relating dataset properties to post-fine-tuning performance, validate these trends at larger model scales and compute budgets, evaluate transfer to natural language benchmarks, and develop formal budget-aware RLVR theory that captures interactions between optimization budget, token limits, and reward sparsity. As shown by \citet{khatri2025art}, this requires significant computational resources.

\bibliography{references}

@book{denis2017space,
  title={Space and spatial cognition: A multidisciplinary perspective},
  author={Denis, Michel},
  year={2017},
  publisher={Routledge}
}

@article{kaplan2020scaling,
  author    = {Jared Kaplan and Sam McCandlish and Tom Henighan and Tom B. Brown and Benjamin Chess and Rewon Child and Scott Gray and Alec Radford and Jeffrey Wu and Dario Amodei},
  title     = {Scaling Laws for Neural Language Models},
  journal   = {arXiv preprint arXiv:2001.08361},
  year      = {2020}
}

@article{hoffmann2022training,
  author    = {Jordan Hoffmann and Sebastian Borgeaud and Arthur Mensch and Elena Buchatskaya and Trevor Cai and Eliza Rutherford and Diego de Las Casas and Lisa Anne Hendricks and Johannes Welbl and Aidan Clark and Tom Hennigan and Eric Noland and Katie Millican and George van den Driessche and Bogdan Damoc and Aurelia Guy and Simon Osindero and Karen Simonyan and Erich Elsen and Jack W. Rae and Oriol Vinyals and Laurent Sifre},
  title     = {Training Compute-Optimal Large Language Models},
  journal   = {arXiv preprint arXiv:2203.15556},
  year      = {2022}
}

@inproceedings{lai2025survey,
  author    = {Hanyu Lai and Xiao Liu and Junjie Gao and Jiale Cheng and Zehan Qi and Yifan Xu and Shuntian Yao and Dan Zhang and Jinhua Du and Zhenyu Hou and Xin Lv and Minlie Huang and Yuxiao Dong and Jie Tang},
  title     = {A Survey of Post-Training Scaling in Large Language Models},
  booktitle = {Proceedings of the 63rd Annual Meeting of the Association for Computational Linguistics (ACL)},
  year      = {2025},
  pages     = {2486--2511}
}

@article{zhang2024scaling,
  author    = {Biao Zhang and Zhongtao Liu and Colin Cherry and Orhan Firat},
  title     = {When Scaling Meets {LLM} Finetuning: The Effect of Data, Model and Finetuning Method},
  journal   = {arXiv preprint arXiv:2402.17193},
  year      = {2024}
}

@article{shen2025exploring,
  author    = {Wei Shen and Guanlin Liu and Zheng Wu and Ruofei Zhu and Qingping Yang and Chao Xin and Yu Yue and Lin Yan},
  title     = {Exploring Data Scaling Trends and Effects in Reinforcement Learning from Human Feedback},
  journal   = {arXiv preprint arXiv:2503.22230},
  year      = {2025}
}

@article{li2024limr,
  author    = {Xuefeng Li and Haoyang Zou and Pengfei Liu},
  title={LIMR: Less is More for RL Scaling}, 
  journal   = {arXiv preprint arXiv:2502.11886},
  year      = {2025}
}

@article{liu2025trust,
  author    = {Xiaoyuan Liu and Tian Liang and Zhiwei He and Jiahao Xu and Wenxuan Wang and Pinjia He and Zhaopeng Tu and Haitao Mi and Dong Yu},
  title     = {Trust, But Verify: A Self-Verification Approach to Reinforcement Learning with Verifiable Rewards},
  journal   = {arXiv preprint arXiv:2505.13445},
  year      = {2025}
}

@article{qwen3,
      title={Qwen3 Technical Report}, 
      author={An Yang and Anfeng Li and Baosong Yang and Beichen Zhang and Binyuan Hui and Bo Zheng and Bowen Yu and Chang Gao and Chengen Huang and Chenxu Lv and Chujie Zheng and Dayiheng Liu and Fan Zhou and Fei Huang and Feng Hu and Hao Ge and Haoran Wei and Huan Lin and Jialong Tang and Jian Yang and Jianhong Tu and Jianwei Zhang and Jianxin Yang and Jiaxi Yang and Jing Zhou and Jingren Zhou and Junyang Lin and Kai Dang and Keqin Bao and Kexin Yang and Le Yu and Lianghao Deng and Mei Li and Mingfeng Xue and Mingze Li and Pei Zhang and Peng Wang and Qin Zhu and Rui Men and Ruize Gao and Shixuan Liu and Shuang Luo and Tianhao Li and Tianyi Tang and Wenbiao Yin and Xingzhang Ren and Xinyu Wang and Xinyu Zhang and Xuancheng Ren and Yang Fan and Yang Su and Yichang Zhang and Yinger Zhang and Yu Wan and Yuqiong Liu and Zekun Wang and Zeyu Cui and Zhenru Zhang and Zhipeng Zhou and Zihan Qiu},
      year={2025},
      journal   = {arXiv preprint arXiv:2505.09388},
}

@inproceedings{hu2021lora,
  author    = {Edward J. Hu and Yelong Shen and Phillip Wallis and Zeyuan Allen-Zhu and Yuanzhi Li and Shean Wang and Lu Wang and Weizhu Chen},
  title     = {{LoRA}: Low-Rank Adaptation of Large Language Models},
  booktitle = {International Conference on Learning Representations (ICLR)},
  year      = {2022}
}

@article{dsouza2025automatingbenchmarkdesign,
  author    = {Amanda Dsouza and Harit Vishwakarma and Zhengyang Qi and Justin Bauer and Derek Pham and Thomas Walshe and Armin Parchami and Frederic Sala and Paroma Varma},
  title     = {Automating Benchmark Design},
  journal   = {arXiv preprint arXiv:2510.25039},
  year      = {2025}
}

@article{guo2025deepseek,
  title={Deepseek-r1: Incentivizing reasoning capability in llms via reinforcement learning},
  author={Guo, Daya and Yang, Dejian and Zhang, Haowei and Song, Junxiao and Zhang, Ruoyu and Xu, Runxin and Zhu, Qihao and Ma, Shirong and Wang, Peiyi and Bi, Xiao and others},
  journal={arXiv preprint arXiv:2501.12948},
  year={2025}
}

@article{shao2024deepseekmath,
  title={DeepSeekMath: Pushing the Limits of Mathematical Reasoning in Open Language Models}, 
  author={Zhihong Shao and Peiyi Wang and Qihao Zhu and Runxin Xu and Junxiao Song and Xiao Bi and Haowei Zhang and Mingchuan Zhang and Y. K. Li and Y. Wu and Daya Guo},
  journal={arXiv preprint arXiv:2402.03300},
  year={2024}
}

@techreport{openai2025gpt5,
  author       = {OpenAI},
  title        = {GPT-5 System Card},
  institution  = {OpenAI},
  year         = {2025},
  month        = aug,
  note         = {Version published August 7. Available at: \url{https://cdn.openai.com/gpt-5-system-card.pdf}}
}

@article{comanici2025gemini,
  title={Gemini 2.5: Pushing the frontier with advanced reasoning, multimodality, long context, and next generation agentic capabilities},
  author={Comanici, Gheorghe and Bieber, Eric and Schaekermann, Mike and Pasupat, Ice and Sachdeva, Noveen and Dhillon, Inderjit and Blistein, Marcel and Ram, Ori and Zhang, Dan and Rosen, Evan and others},
  journal={arXiv preprint arXiv:2507.06261},
  year={2025}
}

@techreport{openai2025o3_n_o4mini,
  author       = {OpenAI},
  title        = {o3 and o4-mini System Card},
  institution  = {OpenAI},
  year         = {2025},
  month        = apr,
  note         = {System card available at: \url{https://deploymentsafety.openai.com/o3/sabotage}}
}

@techreport{anthropic2025claudeSonnet45,
  author       = {Anthropic},
  title        = {Claude Sonnet 4.5 System Card},
  institution  = {Anthropic},
  year         = {2025},
  month        = sep,
  note         = {Available at: \url{https://assets.anthropic.com/m/12f214efcc2f457a/original/Claude-Sonnet-4-5-System-Card.pdf}}
}

@article{zeng2025glm,
  title={Glm-4.5: Agentic, reasoning, and coding (arc) foundation models},
  author={Zeng, Aohan and Lv, Xin and Zheng, Qinkai and Hou, Zhenyu and Chen, Bin and Xie, Chengxing and Wang, Cunxiang and Yin, Da and Zeng, Hao and Zhang, Jiajie and others},
  journal={arXiv preprint arXiv:2508.06471},
  year={2025}
}

@article{wen2025reinforcement,
  title={Reinforcement Learning with Verifiable Rewards Implicitly Incentivizes Correct Reasoning in Base LLMs}, 
  author={Xumeng Wen and Zihan Liu and Shun Zheng and Shengyu Ye and Zhirong Wu and Yang Wang and Zhijian Xu and Xiao Liang and Junjie Li and Ziming Miao and Jiang Bian and Mao Yang},
  journal={arXiv preprint arXiv:2506.14245},
  year={2025}
}

@article{khatri2025art,
  title={The Art of Scaling Reinforcement Learning Compute for LLMs},
  author={Khatri, Devvrit and Madaan, Lovish and Tiwari, Rishabh and Bansal, Rachit and Duvvuri, Sai Surya and Zaheer, Manzil and Dhillon, Inderjit S and Brandfonbrener, David and Agarwal, Rishabh},
  journal={arXiv preprint arXiv:2510.13786},
  year={2025}
}

@article{poddar2024personalizing,
  title={Personalizing reinforcement learning from human feedback with variational preference learning},
  author={Poddar, Sriyash and Wan, Yanming and Ivison, Hamish and Gupta, Abhishek and Jaques, Natasha},
  journal={Advances in Neural Information Processing Systems},
  volume={37},
  pages={52516--52544},
  year={2024}
}

@article{wang2025reinforcement,
  title={Reinforcement learning for reasoning in large language models with one training example},
  author={Wang, Yiping and Yang, Qing and Zeng, Zhiyuan and Ren, Liliang and Liu, Liyuan and Peng, Baolin and Cheng, Hao and He, Xuehai and Wang, Kuan and Gao, Jianfeng and others},
  journal={arXiv preprint arXiv:2504.20571},
  year={2025}
}

@article{tan2025scaling,
      title={Scaling Behaviors of LLM Reinforcement Learning Post-Training: An Empirical Study in Mathematical Reasoning}, 
      author={Zelin Tan and Hejia Geng and Xiaohang Yu and Mulei Zhang and Guancheng Wan and Yifan Zhou and Qiang He and Xiangyuan Xue and Heng Zhou and Yutao Fan and Zhongzhi Li and Zaibin Zhang and Guibin Zhang and Chen Zhang and Zhenfei Yin and Philip Torr and Lei Bai},
      journal={arXiv preprint arXiv:2509.25300},
      year={2025}
}

@article{wang2025tina,
  title={Tina: Tiny reasoning models via lora},
  author={Wang, Shangshang and Asilis, Julian and Akg{\"u}l, {\"O}mer Faruk and Bilgin, Enes Burak and Liu, Ollie and Neiswanger, Willie},
  journal={arXiv preprint arXiv:2504.15777},
  year={2025}
}

@article{he2025deepmath,
  title={Deepmath-103k: A large-scale, challenging, decontaminated, and verifiable mathematical dataset for advancing reasoning},
  author={He, Zhiwei and Liang, Tian and Xu, Jiahao and Liu, Qiuzhi and Chen, Xingyu and Wang, Yue and Song, Linfeng and Yu, Dian and Liang, Zhenwen and Wang, Wenxuan and others},
  journal={arXiv preprint arXiv:2504.11456},
  year={2025}
}

@article{dang2025reinforcement,
  title={Reinforcement Learning for Reasoning in Small LLMs: What Works and What Doesn't},
  author={Dang, Quy-Anh and Ngo, Chris},
  journal={arXiv preprint arXiv:2503.16219},
  year={2025}
}

@article{schulman2025lora,
  author = {John Schulman and Thinking Machines Lab},
  title = {LoRA Without Regret},
  journal = {Thinking Machines Lab: Connectionism},
  year = {2025},
  note = {https://thinkingmachines.ai/blog/lora/},
  doi = {10.64434/tml.20250929},
}
\bibliographystyle{mlsys2025}

\end{document}